\documentclass[12pt]{mythesis}

\usepackage{times}
\usepackage{fullpage}
\usepackage{graphicx}
\usepackage{amsmath}
\usepackage{amsfonts}
\usepackage[square, comma]{natbib}
\usepackage{float}
\usepackage{mathtools}
\usepackage{booktabs}
\usepackage{multirow}

\DeclareMathOperator*{\argmin}{arg\,min}

\usepackage{hyperref}
\usepackage{subcaption}

\begin {document} 
\frontmatter

\pagestyle{empty}
 {\vspace*{0 cm} \hspace*{7.5cm}
\includegraphics[width=55mm,viewport=0 0 235
85,clip]{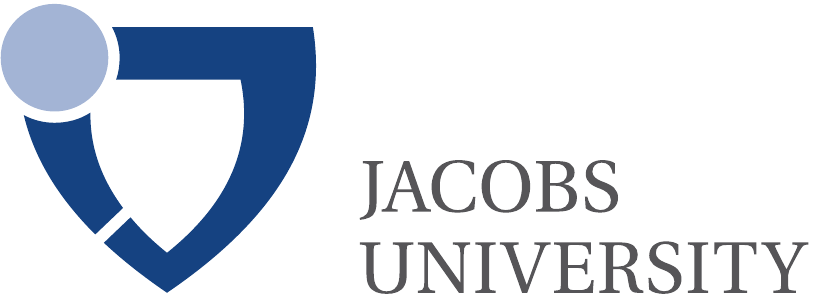}} \vspace{-0.5 cm}

\title{ 
{\bf Harnessing Slow Dynamics in Neuromorphic Computation}}
\author{Tianlin Liu}
\date{\today}
\Year{\the\year}
\trnumber{}

\committee{
Prof. Dr. Herbert Jaeger
}

\coreviewer{
Prof. Dr. Marc-Thorsten H\"utt
}

\support{This research was sponsored by the European Horizon 2020 project NeuRAM3 (grant number 687299), a Jacobs University Graduate Scholarship, and a SMARTSTART1 fellowship provided by the Bernstein Network and the Volkswagen Foundation.}
\disclaimer{}


  
\maketitle



\pagestyle{plain} 

\begin{abstract}

Neuromorphic Computing is a nascent research field in which models and devices are designed to process information by emulating biological neural systems. Thanks to their superior energy efficiency, analog neuromorphic systems are highly promising for embedded, wearable, and implantable systems. However, optimizing neural networks deployed on these systems is challenging. One main challenge is the so-called \emph{timescale mismatch}: Dynamics of analog circuits tend to be too fast to process real-time sensory inputs. In this thesis, we propose a few working solutions to slow down dynamics of on-chip spiking neural networks. We empirically show that, by harnessing slow dynamics, spiking neural networks on analog neuromorphic systems can gain non-trivial performance boosts on a battery of real-time signal processing tasks.

\end{abstract}

\begin{acknowledgments}

I am indebted to many people who have been instrumental in my master level study. First and foremost, my supervisor Professor Herbert Jaeger provided me invaluable guidance, and at the same time, tremendous research freedom. I would like to express my sincere gratitude to him. I would also like to thank my colleagues in the MINDS research group, Fatemeh Hadaeghi and Xu He, for productive collaborations and inspiring scientific discussions. I am grateful to Professor Marc-Thorsten H\"utt for being the co-reviewer of this thesis.

I have greatly benefited from Roberto Cattaneo and Professor Giacomo Indiveri (Institute of Neuroinformatics, Zurich), our NeuRAM3 EU project collaborators. Many research efforts reported in this thesis were initialized based on a fruitful research visit to Zurich hosted by Roberto Cattaneo and Professor Indiveri, without whom this thesis would not have been possible.

I appreciate Jo\~ao Sedoc and Professor Lyle Ungar at the University of Pennsylvania, who hosted me for a summer internship. My internship experience there enabled me to expand my conceptions on various practically relevant machine learning tasks such as natural language processing.

Financially, I gratefully acknowledge the funding of the NeuRAM3 project of the European Horizon 2020 programme, a graduate scholarship provided by Jacobs University, and a SMARTSTART1 fellowship received from the Bernstein Network and the Volkswagen Foundation. 

At a personal level, I thank my parents who unfailingly support me exploring different scientific disciplines throughout my student career. My special thanks go to Jens Pieper, my host-father at Bremen, who provides me with continuous encouragement (together with Br\"otchen, Sinalco, and Kartoffelsuppe) throughout my years of study at Jacobs.

\end{acknowledgments}

\tableofcontents
\listoffigures
\listoftables

\mainmatter


%
%
%
%
%

\chapter{Introduction \label{chap:intro}} 
Computers are ubiquitous in our world, from heavy data crunchers such as supercomputers to wearable devices such as smartwatches. Modern computers have equipped humans with unprecedented ability to process information in ways unimaginable when they were first widely available a few decades ago. Indeed, even today's cell phones have more computational power than computers used in the spaceflight Apollo 11 \citep[Chapter 1]{Kaku2012}, which brought two astronauts to the moon and took them back. With these dazzling advances in computer technologies, we are conditioned to expect that every few years, new generations of computers will always be much faster, smaller, and cheaper. 

Contrary to this accustomed expectation, however, evidence has shown that the computation power of commonly used von Neumann-type computers \citep{vonNeumann1945} will eventually reach a fundamental physical limit. This is due to the combined effects of the imminent end of Moore's law \citep{Kish2002, Cavin2012, Waldrop2016}, the massive energy demands of transistors after the breakdown of Dennard's scaling \citep{Dennard1974}, and the communication bottleneck between central processing units (CPUs) and memory known as the von Neumann bottleneck \citep{Backus1978}. Once these ceilings are reached, the technological advances of today's computers will inevitably flatten out. Hence, there is a compelling need for rethinking computation with paradigms other than the von Neumann architecture.

 Complementary to von Neumann architecture, the neuromorphic computation paradigm \citep{Mead1990} offers one promising route toward designing high-performance and energy-efficient computing devices. Using brain circuits as a source for guidance, neuromorphic systems have a few important advantages when compared to von Neumann systems. Among them, the primary one is the former's superior energy efficiency. Whilst von Neumann systems have CPUs separated from memory components, neuromorphic systems have these elements co-localized. For example, circuit-based synapses on neuromorphic hardware are both the sites for storing memory \emph{and} for performing computation \citep{Indiveri2015, Qiao2017}. These co-localized components effectively decrease the energy consumption induced by memory transfer. For this reason, neuromorphic systems are ideal candidates for wearable devices \citep{Zbrzeski2016}, brain-machine interface modules \citep{Shaikh2019}, speech processing \citep{Braun2019}, mobile robot \citep{Kreiser2018}, and internet of things (IoT) \citep{Gao2019} applications, where low energy consumption is highly desirable \citep{Birmingham2014, Indiveri2015, Furber2016}. 

Despite their notable energy efficiency advantages, most of the neuromorphic devices have not stepped far out of a few pioneering laboratories and industrial research groups. This situation particularly applies to analog neuromorphic hardware, which exhibits a few challenging material properties hindering them from practical applications. These challenging properties include:

\begin{itemize}
\item \emph{Device mismatch}: Due to fabrication imperfections, analog circuits tend to exhibit variabilities and inhomogeneities \citep{Qiao2015}.

\item \emph{Low bit parameter values}: Unlike those of software simulations, programmable parameters of analog neuromorphic hardware usually have bounded ranges, limited resolutions, and low precisions \citep{Chicca2014}. Additionally, oftentimes parameters need to be set globally to a population of neurons but not on the individual neuron level \citep{Moradi2017}.

\item \emph{Timescale mismatch}: Dynamics of analog systems tend to be too fast to process real-time input signals \citep{Chicca2014}.  

\end{itemize}

Among these difficulties, the timescale mismatch problem is particularly troublesome. To be successfully used in application domains such as wearable biosignal monitoring tasks, neuromorphic systems need to have slow timescales which are comparable to those of biological signals. Only in this way can the information contained in input signals be synchronized and integrated using the hardware in real-time. This slow timescale requirement, however, cannot be easily attained with current analog neuromorphic technologies \citep{Chicca2014}. Many neuromorphic systems, therefore, use accelerated timescales. Although these accelerated devices are ideal for simulations that take a very long time in
biological terms \citep{Schemmel2007}, they are unsuitable for real-time signal processing tasks.

Written under the NeuRAM3 EU Horizon 2020 project\footnote{\url{http://www.neuram3.eu/}}, this thesis aims to provide a few working solutions that alleviate the above-mentioned limited timescale problem exhibits by an real-time analog neuromorphic device named Dynap-se \citep{Moradi2017} for real-time signal processing tasks. In this introduction chapter, we first provide an overview of the landscape of neuromorphic hardware. We then review some fundamental computational neuroscience and supervised machine learning notions. Based on these notions, we take an overview of common approaches used to configure neuromorphic devices. The structure, contributions, used sources, and research reproducibility of this thesis are discussed at the end of this chapter.

\section{Neuromorphic computing}

The term neuromorphic computation, coined by Carver Mead \citep{Mead1990}, refers to the use of electronic circuits that emulates biological nervous systems to implement computational mechanisms. In this section, we present a short overview of different types of neuromorphic hardware. 

Neuromorphic systems can be divided into different categories based on different criteria. \citet{Izeboudjen2014} provided an overview of the taxonomies of neuromorphic systems. Despite the varieties of taxonomies, the most common classification criterion is based on the systems' implementation types, i.e., the types of signals processed in circuits. Using this criterion, we can divide neuromorphic hardware into three broad categories: analog, digital, and mixed analog/digital.

Digital neuromorphic hardware share a few characteristics of the ``conventional'' von Neumann-type computers: they use digital transistors to implement boolean-logic gates (such as AND, OR, and NOT), operate with discrete values, and usually employ clocks for synchronization in circuits. Different from conventional computers, however, digital neuromorphic systems are specifically designed to simulate large-scale spiking neural networks by mimicking their biological functionalities. Due to their specializations, circuits on digital neuromorphic hardware consume far lower energy when compared to conventional computers. Additionally, thanks to their digital nature, these neuromorphic systems usually have high precisions and replicable arithmetics, leading to greater user accessibility and fewer computational challenges than analog implementations. However, numerical stabilities of digital neuromorphic systems do not come without a cost: They tend to consume more energy than analog neuromorphic devices \citep{Indiveri2015}. Examples of digital neuromorphic hardware include TrueNorth \citep{Merolla2014}, SpiNNaker \citep{Painkras2012}, and Loihi \citep{Davies2018}. 

 Analog neuromorphic implementation is another variant of neuromorphic hardware. In fact, the term ``neuromorphic,'' when originally defined \citep{Mead1990}, refers to analog neuromorphic systems. These systems use physical characteristics of analog circuits \citep{Andreou1996} to mimic the behaviors of neurons, synapses, and other structures \citep{Liu2002}. To emulate these behaviors, sub-threshold analog circuits require fewer transistors than their digital counterparts \citep{Indiveri2015}.  On-chip spiking neurons on analog neuromorphic hardware are typically asynchronous, acting as independent processors without a central clock. These properties make analog systems closely resemble real biological systems. However, analog systems tend to be noisy, raising challenges for computational algorithms. For example, when sending all neurons in a population constant injection currents, the spiking frequencies of individual neurons tend to vary. \citet{Ning2015} reported 9.4\% variations of spike-frequency variations under the constant injection currents when using the ROLLS processor. Although the variation coefficient 9.4\% is considered to be low when compared to other neuromorphic hardware \citep{Ning2015}, this variability still rules out a large portion of state-of-the-art machine learning algorithms, which are based on floating-point precision operations.  
 
 Analog neuromorphic hardware can be further categorized into two classes: real-time and accelerated \citep[Chapter 1]{Pfeil2015}. In real-time hardware, synapses and neurons operate in timescales similar to their biological counterparts. These systems are usually designed for applications in bio-signal processing, prosthetics, and robotics tasks. In accelerated systems, timescales of the hardware network are usually $10^3$ to $10^4$ faster than their biological counterparts \citep{Indiveri2015}. These systems are suited for applications that take a very long time in
biological terms \citep{Indiveri2015}, e.g., modeling several years of childhood development \citep{Furber2016}. Examples of accelerated analog neuromorphic hardware include Spikey \citep{Briiderle2010} and BrainScaleS \citep{Schemmel2012}. Examples of real-time analog neuromorphic devices include ROLLS \citep{Ning2015} and Dynap-se \citep{Moradi2017}.

Besides digital and analog neuromorphic systems, there exist analog/digital mixed systems. Examples of these systems include Neurogrid \citep{Benjamin2014} and Braindrop \citep{Neckar2019}.

Our working device used in this thesis is Dynap-se \citep{Moradi2017}, an analog and real-time neuromorphic device. We will introduce its features in details in Chapter \ref{chap:neuromorphicComp}.

\section{Recurrent network of spiking neurons}

In the previous section, we have reviewed different types of neuromorphic hardware. Although these types of hardware are designed based on different principles, they all perform computation with on-chip spiking neural networks. To work with neuromorphic hardware, it is therefore necessary to understand a few basic notions related to spiking neural networks. In this section, we briefly review the leaky integrate-and-fire neuron model, which is arguably the simplest form of a neuron model. We then explain how to connect these neurons into a recurrent neural network. Our presentation in this section mainly follows \citep[Chapter 1]{Gerstner2014} and \citep{Nicola2017}.

\subsection{LIF neurons}

The dynamics of a leaky integrate-and-fire (LIF) \citep{Lapicque1907} neuron with index by $i$ at the time $t$ can be formulated by
\begin{eqnarray}
& & \tau_v \frac{d v_i}{d t} =- \left [v_i(t) - v_{\text{rest}} \right ]+R I_i(t) \label{eq:subthr} \\[5pt]
& & \text{If} ~~ v_i(t) > \vartheta, \nonumber \\[5pt]
& & \quad \text{then} ~  v_i(t)  \coloneqq v_{\text{rest}} \label{eq:reset},
\end{eqnarray}
where $v_i$ is the membrane potential of the neuron, $I_i$ is the input current of the neuron, $R$ is the membrane resistance, $\tau_v$ is the membrane time constant, $v_{\text{rest}}$ is the resting potential, and $\vartheta$ is the firing threshold. Equation \ref{eq:subthr} describes the leaky integrator dynamics in the sub-threshold regime of a neuron, i.e., in the time periods between two consecutive spikes. Equation \ref{eq:reset} defines a reset mechanism: Whenever the membrane potential $v_i$ crosses the firing threshold $\vartheta$, $v_i$ is set to be the resting potential $v_{\text{rest}}$.

The spike train produced by the neuron $i$ at the time $t$ can be denoted by

\begin{equation}
s_i(t)=\sum_{t_f^i}\delta(t-t_f^i),
\end{equation}
where $t_f^i$ are the firing times of the neuron $i$ and $\delta(\cdot)$ is a Dirac delta function.

\subsection{Recurrent network of LIF neurons \label{sec:RNN_LIF}}

Following \citep{Nicola2017}, we now formalize how LIF neurons communicate with each other via their spike induced synaptic currents, giving rise to a recurrent neural network (RNN). The dynamics of synaptic currents $r_i$ induced by a spike train $s_i$ of neuron $i$ can be written as
\begin{equation}
 \tau_r \frac{dr_i(t)}{dt}=- r_i(t) + s_i (t),
\end{equation}
where $\tau_r$ is the synaptic time constant.

For a post-synaptic neuron indexed by $i$, each pre-synaptic neuron indexed by $j$ contributes its spike induced synaptic currents $r_j$ to $I_i$. Assuming that these contributions are linear, we write the synaptic currents $I_i(t)$ as

\begin{equation} \label{eq:synapticCurrents}
 I_i(t) \coloneqq \sum_{j} W_{ij} r_j(t) + I_0
\end{equation}
where $W_{ij}$ are real values specifying the magnitude of the spike induced currents arriving at neuron $i$ from neuron $j$ and $I_0$ is a constant current set near or at the rheobase (threshold to spiking) value as used in \citep{Nicola2017}.

Plugging $I_i(t)$ in Equation \ref{eq:synapticCurrents} back to Equation \ref{eq:subthr}, we see the sub-threshold dynamics of the neuron $i$ under the influence of its pre-synaptic neurons can be re-written as

\begin{equation} \label{eq:synaptic2}
 \tau_v \frac{d v_i}{d t} =-\left [v_i(t) - v_{\text{rest}} \right ]+R \sum_{j} W_{ij} r_j(t) + R I_0.
\end{equation}

To take the reset mechanism into account, we add an additional term in Equation \ref{eq:synaptic2} to specify the full dynamics of membrane potential of a LIF neuron:

\begin{equation} \label{eq:synapticCurrentsInteract}
 \tau_v \frac{d v_i}{d t} =-\left [v_i(t) - v_{\text{rest}} \right ]+R \sum_{j} W_{ij} r_j(t)  + R I_0 -  \theta s_i(t).
\end{equation}
where $\theta \coloneqq \vartheta - v_{\text{rest}}$ is the difference between spiking threshold $\vartheta$ and reset potential $v_{\text{rest}}$.

Using more compact matrix notations, assuming that there are $N$ LIF neurons contributing to the recurrent dynamics, we can write the network as

\begin{equation} \label{eq:RNN_LIF_onIn}
\begin{aligned}
  \tau_v \dot{\mathbf{v}} & =-\left [\mathbf{v}(t) - \mathbf{v}_{\mathrm{rest}} \right ]+ R \mathbf{W} \mathbf{r}(t) + R \mathbf{I}_0 - \theta  \mathbf{s}(t), \\
 \tau_r \dot{\mathbf{r}} & = - \mathbf{r}(t) + \mathbf{s} (t),
 \end{aligned}
 \end{equation}
 where $\dot{\mathbf{v}}, \dot{\mathbf{r}}, \mathbf{v}, \mathbf{r}$, and $\mathbf{s}$ are all $N$-dimensional vectors whose $i$-th entries are $\frac{d v_i}{dt}$,  $\frac{d r_i}{dt}$, $v_i$, $r_i$, and $s_i$; $\mathbf{v}_{\text{rest}}$ is a vector with all entries being $v_{\text{rest}}$ and $\mathbf{I}_0$ is a vector with all entries being $I_0$; $\mathbf{W} \in \mathbb{R}^{N \times N}$ is a recurrent connectivity matrix whose $(i, j)$-th entry is $W_{ij}$. 
 
Note that, the network in Equation \ref{eq:RNN_LIF_onIn} is an autonomous system where no external input is defined. To deal with input-driven systems, we assume that at each time $t$, we are given an external input signal taking values as a $m$-dimensional real-valued vector $\mathbf{u}$. With this assumption, we add another term in Equation \ref{eq:RNN_LIF_onIn} to take external driving signal into consideration
 
 \begin{equation} \label{eq:RNN_LIF_pre}
\begin{aligned}
  \tau_v \dot{\mathbf{v}} & =-\left [\mathbf{v}(t) - \mathbf{v}_{\mathrm{rest}} \right ]  + \mathbf{W}^{\text{in}} \mathbf{u} (t) + R \mathbf{W} \mathbf{r}(t)  + R \mathbf{I}_0 - \theta  \mathbf{s}(t),  \\
 \tau_r \dot{\mathbf{r}} & = - \mathbf{r}(t) + \mathbf{s} (t),
 \end{aligned}
 \end{equation}
 where $\mathbf{W}^{\text{in}} \in \mathbb{R}^{N \times m}$ is an input weight matrix. 
 
 To reduce the number of parameters in Equation \ref{eq:RNN_LIF_pre} and make things simpler, we make additional assumption that $v_{\text{rest}} = 0$ and $R = 1$ as done in \citep{Nicola2017} and \citep{Neftci2019}. This reduces the RNN formalism into
 
\begin{equation} \label{eq:RNN_LIF}
\begin{aligned}
  \tau_v \dot{\mathbf{v}} & =- \mathbf{v}(t)  + \mathbf{W}^{\text{in}} \mathbf{u} (t) +  \mathbf{W} \mathbf{r}(t)  + \mathbf{I}_0 - \theta  \mathbf{s}(t),  \\
 \tau_r \dot{\mathbf{r}} & = - \mathbf{r}(t) + \mathbf{s} (t).
 \end{aligned}
 \end{equation}
 
 Equation \ref{eq:RNN_LIF} specifies a RNN with LIF neurons. We remark that, however, this formulation is by no means the only possible version. In fact, most of the existing RNN architectures of LIF neurons (e.g., \citep{Huh2018, Bellec2018, Neftci2019}) use slightly different formalisms. For example, \citet{Neftci2019} use recurrent weights that act on spike trains of pre-synaptic neurons rather than on spike-induced currents of pre-synaptic neurons as we did in Equation \ref{eq:RNN_LIF}. 

\subsection{Supervised training for RNN of LIF neurons}

We now describe how to set up RNNs for supervised, input-output function approximation tasks. For such tasks, oftentimes we are given a collection of time-dependent input signals $\{\mathbf{u}(t)\}_t$ and desired output signals $\{\mathbf{y}(t) \}_t$, where $\mathbf{u}(t) \in \mathbb{R}^m$ and $\mathbf{y}(t) \in \mathbb{R}^k$ for some $m$ and $k \in \mathbb{N}$.  In the training phase, our goal is to configure a RNN such that it produces $\{\mathbf{y}(t)\}_t$ as close as possible (up to some regularization effects) whenever the input signal $\{\mathbf{u}(t)\}_t$ is given. One way to achieve this with our RNN specified in Equation \ref{eq:RNN_LIF} is to invest an additional output matrix $\mathbf{W}^{\text{out}} \in \mathbb{R}^{k \times N}$, such that the following approximation

\begin{equation} \label{eq:solveRNN}
\mathbf{y}(t) \approx \mathbf{W}^{\text{out}} \mathbf{r}(t)
\end{equation}
holds under some metric for all $t$.

To achieve this goal, we need to optimize the parameters $\mathbf{W}^{\text{in}}$, $\mathbf{W}$, and  $\mathbf{W}^{\text{out}}$ under some metrics. The recent standard practice for this optimization task is to use back-propagation-through-time algorithms \citep{Rumelhart1986} with variants of surrogate gradients \citep{Esser2016, Bellec2018, Zenke2018, Shrestha2018}. A recent review for surrogate gradients training methods for spiking neural networks is given by \citet{Neftci2019}. 

Although surrogate gradients training methods for spiking networks have achieved state-of-the-art results with software simulations, when it comes to neuromorphic devices, they may not be applicable for one device or another. In the next section, we provide a brief overview of the applicability of learning algorithms of spiking neural networks for neuromorphic devices.

\section{Learning algorithms for neuromorphic computation}

 We have already introduced neuromorphic hardware as well as spiking neural networks as a computation paradigm deployable to neuromorphic hardware. In this section, we consider the strategies for optimizing parameters of neural networks on neuromorphic hardware. 

 Since different types of neuromorphic hardware have different constraints, the choice of learning algorithms for on-chip neural networks heavily depends on the device one uses. By and large, learning algorithms for neural networks on neuromorphic hardware are mainly advancing along two lines of investigations \citep{He2019}: a deep learning \citep{Goodfellow2016} approach and a reservoir computing \citep{Jaeger2001, Maass2002} approach.
 
 \subsection{Deep learning for neuromorphic hardware \label{sec:deepLearningNeuromorphic}}
 
 As deep neural networks (DNNs) have achieved highly remarkable results on important machine learning tasks such as image classification \citep{He2016}, machine translation \citep{Bahdanau2015}, and speech processing \citep{Amodei2016}, numerous studies are devoted to transferring the success of conventional-computer-based deep learning algorithms to their neuromorphic hardware counterparts. As observed by \citet{Liu2018b}, most research in this line of investigation leverages a pre-training approach. That is, one first trains a DNN of artificial neurons or spiking neurons on a conventional computer with standard techniques and then maps the trained parameters to neuromorphic hardware. Since the parameter mapping needs relatively high precision, most of the work in this approach uses digital hardware as the neuromorphic platform. For example, \citet{Jin2010} and \citet{Stromatias2015} use SpiNNaker; \citep{Esser2015,Esser2016} use TrueNorth. More recently, \citet{Schmitt2017} show that a similar approach works for BrainScaleS analog neuromorphic system. The idea is to first roughly map the parameters estimated from DNN to BrainScaleS hardware, and then iteratively fine-tune the parameters in a ``hardware in the loop'' fashion. This is realized with the help of an interface between the conventional computer and the BrainScaleS hardware. 
 
Although the deep learning paradigm has been empirically proven to be highly successful for many neuromorphic devices, there are a few reasons why it is not immediately suitable for our Dyanp-se hardware. For one, the learned parameters of DNN cannot be mapped to Dynap-se conveniently as the hardware only has limited parameter resolution. Additionally, the variability of on-chip neurons may cripple the mapped DNN architecture since the performance of DNN relies on highly precise and well-orchestrated parameters. What is more, a hardware-in-the-loop method similar to \cite{Schmitt2017} cannot be realized easily on Dynap-se\footnote{That being said, a recently released front-end interface of Dynap-se named CortexControl (\url{https://ai-ctx.gitlab.io/ctxctl/primer.html}) brings some promises to this approach.}.  
  
  \subsection{Reservoir computing for neuromorphic hardware}
The Reservoir Computing paradigm \citep{Jaeger2001, Maass2002} offers a second route for training recurrent spiking neural networks on neuromorphic systems.  Concretely, the reservoir computing paradigm is usually realized with the following steps \citep{Jaeger2007}. We introduce these steps by using our RNN of Equation \ref{eq:RNN_LIF} as a concrete example.  
 
 \begin{enumerate}
 \item Set up a random RNN. In our example of RNN of LIF neurons specified in Equation \ref{eq:RNN_LIF}, this amounts to randomly create $\mathbf{W}^{\text{in}}$ and $\mathbf{W}$ up to some hyperparameters which govern the randomness of these matrices.
 
 \item Drive the RNN with input signals to harvest reservoir states, i.e., temporal features produced by recurrent neurons. In our RNN of LIF neuron example, this can be practically done by choosing a sequence of discretized time $\{t_k\}$ and collect $\mathbf{s}(t_k)$ for all $t_k$ by using Equation \ref{eq:RNN_LIF}. The collected spike train $\{\mathbf{s}(t_k\})$ can be further smoothed into $\{\mathbf{r}(t_k)\}$ by using an exponentially decay filter specified in Equation \ref{eq:RNN_LIF}. Those $\{\mathbf{r}(t_k)\}$ can be seen as high-dimensional features of the input signal $\{\mathbf{u}(t_k)\}$.
 
  \item Read out the desired outputs by linearly combining the reservoir states. In our RNN example, we can estimate an output matrix $\mathbf{W}^{\mathrm{out}}$ which linearly combines reservoir states $\mathbf{r}(t_k)$ into the desired target signal $\mathbf{y}(t_k)$ for all $t_k$. A commonly used approach to realize this is to solve Equation \ref{eq:solveRNN} via a ridge regression
\begin{equation} \label{eq:reservoirRegression}
\mathbf{W}^{\mathrm{out}}= \mathbf{Y} \mathbf{\Phi}^\top \left(\mathbf{\Phi} \mathbf{\Phi}^\top +\alpha \mathbf{I}\right)^{-1},
\end{equation} 
where $\alpha$ is a Tikhonov regularization coefficient, $\mathbf{\Phi}$ is a matrix whose columns are $\mathbf{r}(t_k)$, $\mathbf{Y}$ is a matrix whose columns are those target $\mathbf{y}(t_k)$, and $\mathbf{I}$ is an identity matrix \citep{Lukosevicius2012}.
 \end{enumerate}
 
 Unlike the deep-learning based pre-training approach, the reservoir computing approach is usually directly carried out using neuromorphic hardware. Compared to DNNs, the number of parameters needed to be estimated for the reservoir computing approach is much smaller: The recurrent weights $\mathbf{W}^{\text{in}}$ and $\mathbf{W}$ are fixed throughout the training and testing phase; only $\mathbf{W}^{\text{out}}$ needs to be estimated. This greatly simplifies the optimization procedure. More importantly, since $\mathbf{W}^{\text{in}}$ and $\mathbf{W}$ are random matrices, the inherent variability of on-chip neurons of analog hardware can be seen as an advantage rather than a shortcoming for deploying the reservoir computing pipeline. For this reason, reservoir computing has been perceived as a suitable paradigm for analog neuromorphic computation.
 
\section{Thesis overview}

So far we have reviewed various notions related to neuromorphic computation. Building upon these notions, this thesis is structured as follows. Chapter \ref{chap:neuromorphicComp} gives an overview of our neuromorphic hardware, the Dynap-se board. We provide a general routine for performing experiments on Dynap-se. Several issues related to the practical implementations of the routine are discussed. 

Chapter \ref{chap:paramtuning} presents two parameter selection heuristics that we empirically found to be useful. By selecting a few time constants for ordinary differential equations which characterize the dynamics of non-chip neurons as well as the synapse types of neurons, we nudge the on-chip neural network toward having a slower timescale. We conducted a few synthetic experiments to probe the dynamics of on-chip neural networks. These experiments show that the heuristically tuned parameters yield slower neural dynamics when compared to untuned ones. 

Chapter \ref{chap:resTransfer} introduces the reservoir transfer paradigm. This scheme ``mirrors'' the dynamic properties of a well-performing artificial recurrent network (optimized on a conventional computer) to spiking recurrent networks deployed on a Dynap-se neuromorphic microchip. We conducted experiments using ECG heartbeat classification tasks to test the proposed method. For the ECG classification task, the empirical performance achieved by Dynap-se hardware favorably approaches the performance achieved by software simulations. 

We conclude this thesis with Chapter \ref{chap:conclusion}. Limitations of the current work are summarized and a few lines of future investigations are outlined.  

\section{Used sources}
This thesis partially uses results reported previously. Some parts of Chapter \ref{chap:neuromorphicComp} and Chapter \ref{chap:paramtuning} are from my independent study report \citep{Liu2018} completed in Spring 2018. Chapter \ref{chap:resTransfer} is an extended version of the paper \citep{He2019} and the contributions of the authors are documented at the beginning of the Chapter \ref{chap:resTransfer}. 

In numerical experiments of this thesis, we use DYNAPSETools\footnote{\url{https://sanfans.github.io/DYNAPSETools}} software package. Developed by \citet{Cattaneo2018}, the software package is a collection of python classes and modules for the purpose of processing spike events produced by Dynap-se.

\section{Research reproducibility}

The code for replicating numerical experiments reported in Chapter \ref{chap:paramtuning} and Chapter \ref{chap:resTransfer} together with their respective used data collected from Dynap-se are available on the GitHub\footnote{\url{https://github.com/liutianlin0121/msc_thesis_code}}.

\chapter{Dynap-se Neuromorphic Microchips \label{chap:neuromorphicComp}} 
In this chapter, we introduce the Dynap-se hardware  \citep{Moradi2017}, which is our working device used throughout this thesis. Dynap-se is the acronym for \textbf{Dy}namic \textbf{N}euromorphic \textbf{A}synchronous \textbf{P}rocessor in a \textbf{S}calabl\textbf{e} variant. The name indicates that the hardware is able to perform computations in an asynchronous fashion and is scalable to large neural network architectures. In this chapter, we first introduce the general feature of Dynap-se hardware. We then provide a pipeline for conducting experiments using Dynap-se. Last, we describe how do we concretely implement this pipeline.

\section{Dynap-se board} 

\begin{figure}[H]
  \centering
  \includegraphics[width= 0.6 \textwidth ]{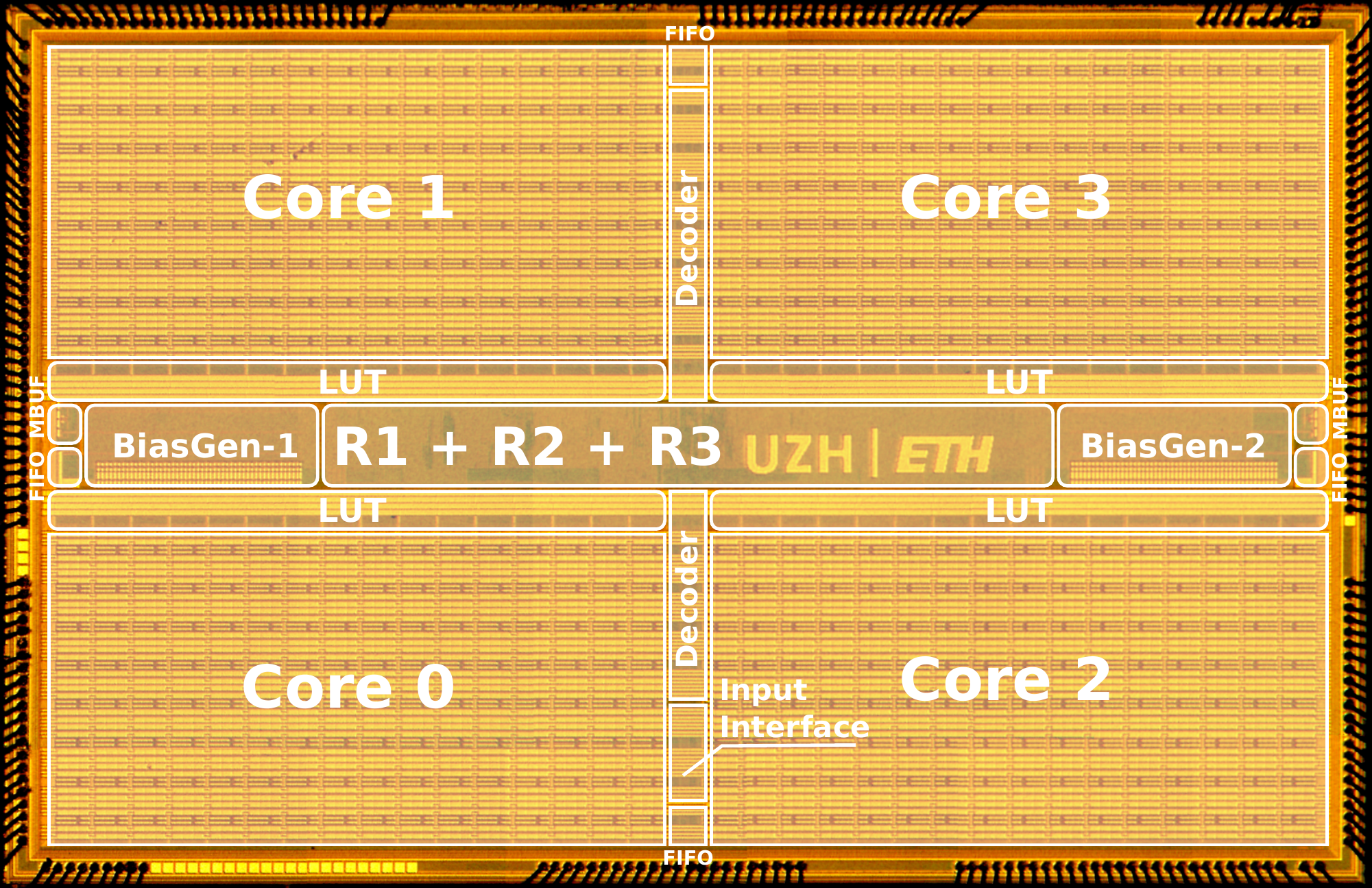}
  \caption{The multi-score structure of a Dynap-se microchip \citep{Moradi2017}. \label{fig:cores}}
\end{figure}

The Dynap-se board that we are using contains four chips, each chip mainly contains four interconnected blocks, which are called cores. The schematic layout of these four cores (Core 0 to Core 3) is shown in Figure \ref{fig:cores}. Each core in a chip contains 256 neurons. 

Besides four main blocks, Core 0 to Core 3, there are other blocks such as BiasGen-1, BiasGen-2, R1, R2, and R3 as shown in Figure \ref{fig:cores}. These blocks are placed to govern the on-chip neural dynamics, e.g., set up connectivity topologies, neuron parameters, and synapse parameters. 

On a conventional computer, Dynap-se can be configured with the support of cAER\footnote{ \url{https://inivation.com/support/software/caer/}}, which is an open-source event-based processing framework written in  C and C++. The cAER framework provides a collection of modules for configuring and monitoring on-chip neural networks. It has a convenient graphical user interface\footnote{\url{https://github.com/inivation/caerctl-gui-javafx}} (GUI). Among others, the functionalities of the GUI of cAER include (i) setting parameters for on-chip neurons, (ii) loading neural network architectures, (iii) sending input spike-based stimuli, and (iv) recording the output spike events. The functionalities of these GUI-based operations will be introduced in our summarized experiment pipeline in Section \ref{sec:pipeline}. 

\subsection{On-chip neurons}

The on-chip neurons implemented on Dynap-se are designed to emulate neurons of Adaptive Exponential Integrate-and-Fire (AdEx) model \citep{Brette2005}, which is a generalization of the leaky integrate-and-fire model. Properties of on-chip neurons can be tuned by a programmable bias-generator, which contains 25 parameters such as injection current level, refractory period length, time constants, and synaptic efficacy. A detailed list of these parameters can be found in the Dynap-se user guide \citep{DynapseUserguide}. The values of these parameters have low-bit resolutions. As an example, the refractory period of a neuron can only be specified as a tuple of coarse and fine values, where a coarse value can be chosen as an integer from 0 to 7, and a fine value can be chosen as an integer from 0 to 255. In addition, these parameters can only be set globally for each core but not for an individual neuron. Due to device mismatch, effective values of these parameters may vary across different neurons. As a result, although all neurons within a core share the same parameter values, every individual neuron exhibits different behavior \citep{DynapseUserguide}. In addition, only neurons' spike trains can be recorded by Dynap-se. Neurons' state variables such as currents and membrane potentials, however, cannot be recorded.



\subsection{On-chip neural networks} \label{sec:onChipNN}

So far we have introduced the dynamics of single neurons on Dynap-se. For computational tasks, however, oftentimes we wish to connect individual neurons into a neural network on Dynap-se. To define a topology (connectivity pattern) of an on-chip neural network, we need to use the NetParser module of cAER\footnote{\url{https://github.com/inivation/caer}} to specify the connections. To understand the workflow of configuring an on-chip neural network, we first need to explain the difference between ``virtual'' and ``real'' neurons on Dynap-se.

To process a sequence of input spike train with a population of neurons, we first send this spike train to its designated receivers in the neuron population. Conceptually, these input spikes can be seen as the neuronal responses produced by some external neurons which will not participate in recurrent connections once their produced spikes leave them. In Dynap-se, such source neurons are referred to as ``virtual neurons''. A virtual neuron cannot send spikes to another virtual neuron, reflecting their ``input'' nature.  

We can use such virtual neurons to send input spikes to ``real neurons,'' which are neurons that can communicate with each other via synaptic connections. After the real neurons receive the input spikes from virtual neurons, they will process the spikes and potentially propagate newly generated spikes to other real neurons with which they connect, depending on the network topology. For each synaptic connection, we can specify the connection efficacy and synapse type. The efficacy of a synaptic connection needs to be defined in terms of content-addressable memory (CAM). For each neuron, 64 CAMs in total are allowed for fan-in and fan-out connections. Each synapse can be realized with four connection types: slow inhibitory, fast inhibitory, slow excitatory, and fast excitatory. Excitatory synapses increase the membrane potential of postsynaptic neurons while inhibitory synapses lower membrane potential of postsynaptic neurons. ``Fast'' synapses on Dynap-se emulate synapses with AMPA receptors, while ``slow'' synapses on Dynap-se emulate synapses with NMDA receptors. These synapses are called ``fast'' and ``slow'' because one key difference between synapses with NMDA receptor and those with AMPA receptors is that the former enable membrane potential to have slower onsets and have decays that last longer \citep[Chapter 5]{Nestler2008} than the latter. 

With the network topology, synapse efficacies, and synapse type chosen, we are able to configure on-chip neural networks by uploading a \texttt{.txt} file in the NetParser module. The specific format of this \texttt{.txt} file will be introduced in Section \ref{sec:pipeline}. 

\section{Conducting numerical experiments on Dynap-se} \label{sec:pipeline}

In this section, we take a technical overview of the general routine of using Dynap-se for computation. We then introduce our working solutions for a few key steps in the routine. 
 
 \subsection{A general routine for performing numerical experiments}
 
\begin{description}
 \item[Step 1: Define the input spike train.]  \hfill \vspace{0.05 in} \\
  To start the experiment, one needs to determine the input patterns.  The input pattern might be continuous digital signals or discontinuous spike trains. If the input signals are continuous (e.g., sine waves), they have to be converted into spikes first via a spike-encoding mechanism. 
  

\item[Step 2: Write spikes into a Dynap-se readable format] \hfill \vspace{0.05 in} \\ 
With the input spike data, we proceed to define the sender (source neuron) and receiver (target neuron) of the input spikes. As we have introduced earlier, the senders of such input spikes are virtual neurons. To send spikes from virtual neurons to real neurons, one needs to specify 2 numbers. The first number is the sender-receiver correspondence, which is a number encoded by three variables: (i) the virtual neuron ID, (ii) the virtual chip ID, and (iii) the destination core(s); the second number is the waiting time between the previous spike and the current spike in the unit of 90 ISI-Bases, where one ISI-Base is 1/90 Mhz = 11.11 nanoseconds. The first number ``sender-receiver correspondence'' deserves more explanations. To encode these three variables, we first convert them individually into a binary number, then concatenate into a long string, and finally convert the string of binary numbers back into to a single decimal number. For a concrete example, suppose we want to send a spike from the 21st neuron (virtual neuron ID = \texttt{20}) on the first virtual chip (virtual chip ID = \texttt{00}) to all of the 4 cores of chip 0 that contain real neurons. The coding mechanism works as follows. First consider the virtual neuron ID -- it is \texttt{10100} because \texttt{10100} is the binary conversion of \texttt{20}; next consider the virtual chip ID -- it is just \texttt{00}; third consider the receiver -- they are cores 0, 1, 2, and 3, so they can be hot coded into \texttt{1111}, where each 1 is an indication that one core has been selected. Putting these 3 variables together, we have \texttt{10100001111}, which will be treated as a binary number and will be converted into a decimal number 1295. The number 1295 is the sender-receiver correspondence. Note that the receivers are not individual neurons, but all neurons in one core or multiple cores. 

The final output of this step is a list of pairs  $(E_0,T_0), (E_1,T_1), \cdots, (E_N,T_N) $, where each $E_i$ for $i \in \{0, \cdots, N\}$ and $N \in \mathbb{N}$ is a sender-receiver correspondence and each $T_i$ for $i \in \{0, \cdots, N\}$ and $N \in \mathbb{N}$ is the waiting time in the unit of 90 ISI-Bases. Such a list should be written into a \texttt{.txt} file, one pair per line, such that they can be fed into Dynap-se using the FPGA-SpikeGen module in the GUI of cAER.

\item [Step 3: Choose neural network parameters] \hfill \vspace{0.05 in} \\
Having the input spike trains written in a Dynap-se readable format, we are ready to send them into Dynap-se. Before doing that, however, we need to specify the parameters of the on-chip neural network. Such parameters include neuron parameters, synapse parameters, and network topology. While neuron parameters and synapse parameters can be easily specified by using the GUI of cAER, the configuration of network topology needs more explanation. For synapse that connects two neurons, we need to provide four pieces of information in the \texttt{.txt} file: (i) the pre-synaptic neuron address, (ii) the connection type, (iii) the CAM slots, and (iv) the post-synaptic neuron address. An address for a pre-synaptic or post-synaptic neuron has three elements: a chip ID, a core ID, and a neuron ID. For example, \texttt{U00-C01-N002} is the address of the neuron 2 of core 1 of chip 0. Dynap-se contains four connection-type, slow inhibitory, fast inhibitory, slow excitatory, and fast excitatory, which are coded by numbers 0, 1, 2, and 3 respectively. The values of CAM slots can be chosen from 1 to 64. As a concrete example, suppose we wish to connect a pre-synaptic neuron, which is the neuron 2 of core 1 of chip 0, to a post-synaptic neuron, which is the neuron 4 of core 3 of chip 2 with a slow inhibitory synapse taking 5 CAMs, we need to write 
\[ \underbrace{\texttt{U00-C01-N002}}_{\text{pre-synaptic neuron ID}} \texttt{->} \underbrace{\texttt{0}}_{\substack{\text{synapse} \\ \text{type}}} \texttt{-} \underbrace{\texttt{5}}_{\substack{ \text{CAM} \\ \text{slots}}} \texttt{-} \underbrace{\texttt{U02-C03-N004}}_{\text{post-synaptic neuron ID}} \] 
To configure a network, a list of these connectivities needs to be provided.

 \item [Step 4: Send input spikes and collect output spikes] \hfill \vspace{0.05 in} \\
 Having the neural network model ready in the previous step, in this step, we send input spikes and collect output spikes using the GUI of cAER software. We first read the \texttt{.txt} file for input spikes and send it into Dynap-se. Next, we collect the output spike-events, which are in the format of \textbf{A}ddress \textbf{E}vent \textbf{DAT}a (AEDAT)\footnote{\url{https://inivation.com/support/software/fileformat/\#aedat-3}}.
 
 \item [Step 5: Use the output spikes for neural network training] \hfill \vspace{0.05 in} \\
 With the collected output spikes, we can visualize and analyze them on a digital computer. A usual recipe is to first post-process the collected spikes into continuous-valued signals and then perform pattern classification or regression tasks using the smoothed spike data. Our collaborators in Zurich have developed a collection of spike-events processing programs\footnote{\url{https://github.com/sanfans/DYNAPSETools}}, which provides a convenient interface for analyzing spike data collected from Dynap-se.

\end{description}
 
 \subsection{Practical implementation of the routine}
 
We have already summarized a general pipeline for conducting experiments using Dynap-se. Yet, to realize this pipeline, we need to be more concrete at each step. Here we spell out a few working solutions we used in our experiments. 

In Step 1 of the experiment routine, sometimes we need to convert continuous signals to input spike trains. Throughout this work, we use a simple method to do the signal-to-spike conversion: If the increase/decrease of a signal relative to the signal value corresponding to the time of its previous spike is above a certain threshold, a spike is placed. We chose this conversion method mainly due to its simplicity. There exists more sophisticated methods (e.g., \citep{Schrauwen2003} and \citep[Chapter 2]{Eliasmith2004}).

The neural network parameters introduced in Step 3 are also subjected to users' choice. Since neuron parameters and network topologies are the main components of learning and adaptation in neural networks, it is not surprising that different choices of parameters will influence the optimality of experiment outcomes. Chapter \ref{chap:paramtuning} and \ref{chap:resTransfer} will be devoted to explaining our working solutions to choose neuron parameters and network topologies. 

Another subjective choice occurs in Step 5 of the experiment routine. To post-process the collected spikes into continuous signals, throughout this work,  we convolve the spikes with an exponential decay kernel. That is, we add exponential tails to all spikes.

\chapter{Slowing down Neuronal Dynamics by Modifying Properties of Individual Neurons \label{chap:paramtuning}} 
In the previous chapter, we have introduced our Dynap-se device. For practitioners, Dynap-se can be seen as an input-output device characterized by tunable parameters and on-chip neural network architectures, producing output spikes whenever input spikes are given. The produced spike representations can then be used for tasks such as pattern recognition. However, configuring Dynap-se to produce practically useful spike representations is challenging due to its material properties such as low bit resolution of tuning parameters, unobservable state variables, device mismatch, and timescale mismatch. Amongst these challenges, the timescale mismatch issue is prominent: The dynamics of on-chip neurons tend to be too fast to maintain relatively long memory spans. In this chapter, we provide a few working solutions to alleviate this problem. Concretely, we offer a few heuristics for tuning neuron and synapse parameters, which nudge the neural networks toward having slower dynamics. We examine the neuronal dynamics characterized by the tuned parameters with three numerical experiments: A \texttt{Pulse} experiment for reservoir visualization,  a \texttt{Chirp} regression task, and a \texttt{Ramp + Sine} pattern classification task.


\section{Heuristics of parameter selection}



To configure time constants that govern the dynamics of on-chip neurons, we study the Differential Pair Integrator (DPI) circuits of Dynap-se, which are circuits that simulate synapses of neurons \citep{Chicca2014}. In essence, the response of a DPI can be modeled by a first-order linear differential equation \citep{Chicca2014}

\[\tau  \frac{d}{dt} I_{\text{out}} + I_{\text{out}} = \frac{I_{\text{th}}}{I_{\tau}} I_{\text{in}}, \]
where $I_{\text{out}}$ is the output of the circuits, i.e., the postsynaptic current of a neuron, $I_{\text{in}}$ is the input current to the synapse,  $I_{\text{th}}$ is a time constant, and $\tau \coloneqq C \frac{U_T}{\kappa I_{\tau}}$ is another time constant for $C$ being the circuit capacitance, $U_T$ being the thermal voltage \citep[Chapter 2]{Liu2002}, $\kappa$ being the subthreshold slope factor, and $I_{\tau}$ being a tunable constant. 

To slow down the dynamics of $I_{\text{out}}$ given $I_{\text{in}}$, we aim to make $\left (\frac{d}{dt} I_{\text{out}} \right )^2$ as small as possible. This can be done by adjusting $ I_{\tau}$ and $I_{\text{th}}$ as tunable parameters and treat other parameters as fixed constants. With some linear algebraic operations, we see 

\begin{eqnarray}
\left (\frac{d}{dt} I_{\text{out}} \right )^2 & = & \left [ \frac{1}{\tau} ( \frac{I_{\text{th}}}{ I_{\tau}} I_{\text{in}} - I_{\text{out}}) \right ]^2 \nonumber  \\
& = &  \left [ \frac{\kappa I_{\tau}}{C U_T} ( \frac{I_{\text{th}}}{I_{\tau}}I_{\text{in}} - I_{\text{out}}) \right ]^2 \nonumber \\
& = & \left [ \frac{\kappa }{C U_T} ( I_{\text{th}} I_{\text{in}} - I_{\text{out}} I_{\tau} ) \right ]^2 \label{eq:1} \label{eq:dpi}.
\end{eqnarray}

To minimize $(\frac{d}{dt} I_{\text{out}})^2$ for fast synapses, Equation \ref{eq:dpi} motivates us to set $I_{\text{th}}$ and $I_{\tau}$ to be the smallest possible values on Dynap-se. In cAER software of Dynap-se, this is done by assigning coarse and fine value of each parameter \texttt{NDPDPIE\_THR\_F\_P} (which characterizes $I_{\text{th}}$ for fast excitatory neurons), \texttt{NDPDPII\_THR\_F\_P} (which characterizes $I_{\text{th}}$ for fast inhibitory neurons), \texttt{DPDPIE\_TAU\_F\_P} (which characterizes $I_{\tau}$ for fast excitatory neurons)
, and \texttt{NDPDPII\_TAU\_F\_P} (which characterizes $I_{\tau}$ for fast inhibitory neurons) to be 0 and 7. \\ 

\fbox{
  \parbox{0.9 \textwidth}{
   \vspace{0.05 in}
   \center
    \textbf{Heuristic 1} \\
     Set the coarse and fine value of \texttt{DPDPIE\_THR\_F\_P} to be 7 and 0, respectively. \\
    Do the same setting for \texttt{NDPDPII\_THR\_F\_P}, \texttt{DPDPIE\_TAU\_F\_P}, and \texttt{DPDPII\_TAU\_F\_P}.
    \vspace{0.05 in}
  }
} \\
    \vspace{0.2 in}

We now introduce the second heuristic, which is about how to specify types of neuron synapses. Intuitively, for the recurrent connections, we want the population of recurrent neurons to act as a memory buffer, such that the characteristics of input signals will be slowly washed out over time.  Recall from Section \ref{sec:onChipNN} that, on Dynap-se, slow synapses emulate biological synapses which enable membrane potential to have slow onsets and long decays. For this reason, we chose slow synapses for reservoir neurons. Concretely, in the NetParser module of cAER, we choose the connection-type IDs of synapses connecting pairs of recurrent neurons to be 0 or 2, which correspond to slow inhibitory and slow excitatory synapses. We set fast synapses for input connections, by choosing the connection-type IDs of synapses between input (virtual) neurons and recurrent neurons to be 1 or 3 using NetParser module of cAER. \\

\fbox{
  \parbox{0.9 \textwidth}{
   \vspace{0.05 in}
   \center
    \textbf{Heuristic 2} \\
     Use fast synapses for input connections. \\
    Use slow synapses for recurrent (reservoir) connections.
    \vspace{0.05 in}
  }
} 
    \vspace{0.2 in}
\\

With these parameters of Dynap-se tuned based on these two heuristics, we now proceed to test the effects of the tuned parameters.

\section{Numerical experiments}

We aim to probe the dynamics of on-chip neurons which are characterized by different parameters via numerical experiments. To evaluate the performance of our tuned parameters, we need to set up the experiments such that the tuned parameters can be fairly compared to untuned ones. 
\subsection{Experiment setup: baseline reservoir and tuned reservoir}

To examine whether the tuned parameters slow down the dynamics of neurons, we define a baseline reservoir and a tuned reservoir, which share the same network topology but differ by neuron and synapse parameters. This shared network topology for both baseline and tuned reservoirs is described in more details below. 

\paragraph{The shared reservoir topology}
The shared network topology we employed here is a topology provided by Roberto Cattaneo, one of our main project collaborators in Zurich. This topology loosely follows the one specified in \citep[Appendix B]{Maass2002}. More specifically, the reservoir takes form as a population of 256 neurons, among which 80\% are excitatory neurons and 20\% are inhibitory neurons, chosen randomly.  By ``excitatory neuron'' or ``inhibitory neuron,'' we mean that these neurons make excitatory or inhibitory synaptic connections with all their respective post-synaptic neurons. We can index all neurons in the reservoir by their respective coordinates in the set $\{(x, y) \} \coloneqq \{0, \cdots, 15 \} \times \{0, \cdots, 15 \}$, where $\times$ denotes the cartesian product. The connectivity structure is defined as follows. For a fixed excitatory neuron with coordinate $(\tilde{x}, \tilde{y})$ and for an arbitrary neuron with coordinate $(x, y)$, the probability of existing a synaptic connection between neuron with coordinate $(\tilde{x}, \tilde{y})$ and neuron with coordinate $(x, y)$ is $\min \left ( C_{\text{exi}}  \exp(-\frac{(\tilde{x} -x)^2 + (\tilde{y} - y)^2}{(2 \lambda_{\text{exi}}^2)}), 1 \right )$, where $C_{\text{exi}} = 0.3$ and $\lambda_{\text{exi}} = 2$.  Similarly, for a fixed inhibitary neuron with coordinate $( \hat{x}, \hat{y})$ and for an arbitrary neuron with coordinate $(x, y)$, the connectivity probability is $\min \left (C_{\text{inh}}  \exp(-\frac{(\hat{x} -x)^2 + (\hat{y} - y)^2}{(2 \lambda_{\text{inh}}^2)}), 1 \right )$, where $C_{\text{inh}} = \lambda_{\text{inh}} = 2$. 

We provide some remarks for this topology. Note that, for a fixed pre-synapse neuron, its connection with a post-synapses neuron only depends on the \emph{coordinate} of the post-synapses neuron, and independent of the \emph{neuron type} (excitatory or inhibitory) of the post-synapses neuron. This implementation is consistent with Dale's principle \citep{Eccles1954}, which states that all synapses originating from the same presynaptic neuron perform the same chemical action at all of its postsynaptic neurons, regardless of the identity of the postsynaptic neuron. However, we notice that this implementation is not the same as what has been proposed in \citep[Appendix B]{Maass2002}, where different connection probabilities are assigned to excitatory-to-excitatory,  excitatory-to-inhibitory, inhibitory-to-excitatory, and inhibitory-to-inhibitory neuronal connectivities.  

\paragraph{Baseline reservoir} The neurons in the default reservoir are characterized by the parameters listed in Table \ref{tab:defaultParams} in the Appendix, which can be configured by pressing the ``set default bias'' button on the netParser interface of Dynap-se. As done in \citep{Cattaneo2018}, all neurons in the baseline reservoir are set to be fast neurons. That is, all neurons make fast synaptic connections with their respective post-synaptic neurons. 



\paragraph{Tuned reservoir} The neurons in the tuned reservoir are characterized by parameters modified according to Heuristic 1 given in the previous section. The full list of tuned parameters can be found in Table \ref{tab:tunedParams} in the Appendix. In addition, all neurons in this reservoir are set to be slow neurons according to the recommendation of Heuristic 2 given in the previous section. That is, all neurons make slow synaptic connections with their respective post-synaptic neurons. 

\subsection{The \texttt{Pulse} experiment \label{subsec:pulse-chirp}}

\begin{figure}[H]
        \centering
            \includegraphics[width=\textwidth]{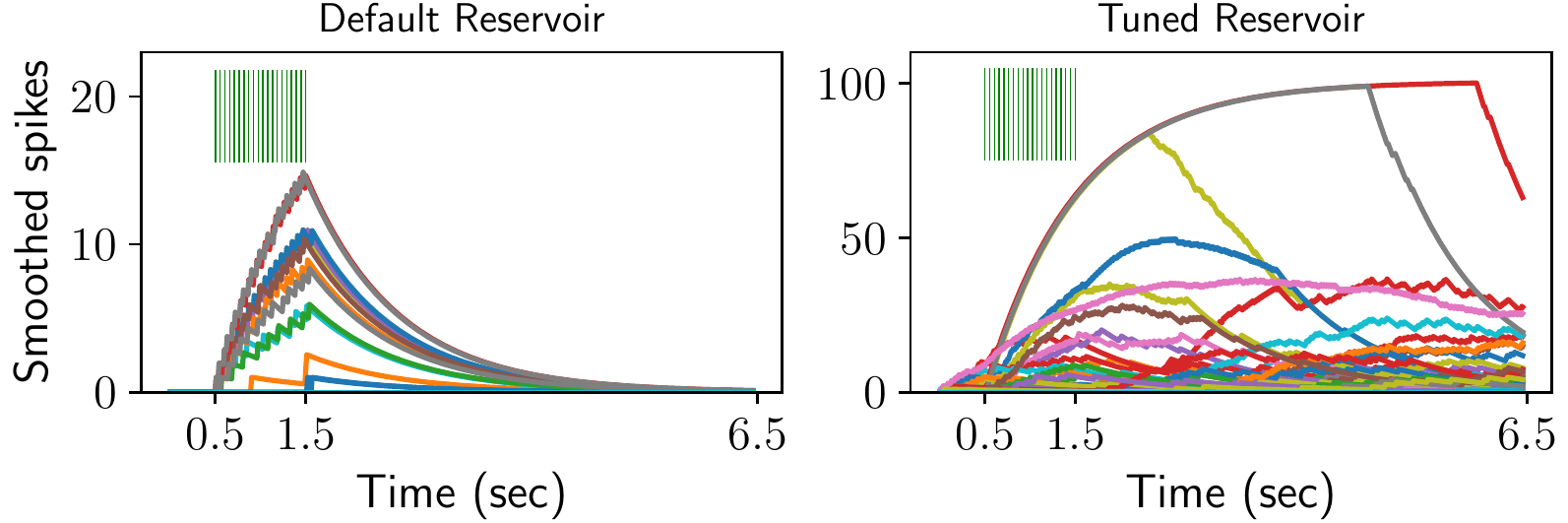}
          \vspace{-0.2 in}
     \caption{Visualization of the reservoir responses driven by a \texttt{Pulse} input. The \texttt{Pulse} input signals are visually illustrated with green vertical bars. For each of the default and tuned reservoir, we randomly choose 100 neurons and plot their neuronal responses (exponentially smoothed spikes) against time. Left: responses of neurons from the default reservoir. Right: responses of neurons from the tuned reservoir.
      \label{fig:reservoirResponse_pulse}}    
    \end{figure}

In this experiment, we aim to visualize and examine the reservoir responses driven by simple driving signals. To this end, we used a pulse of spikes as input to drive the reservoir. The experiment lasts for 6.5 seconds. For the initial 0.5 seconds and last 5 seconds, there is no spike; from 0.5 seconds to the 1.5 seconds, we sent a sequence of equally spaced spikes, where the distance between two nearby spikes was fixed to be 0.001 seconds. After post-processing the spike trains produced by reservoir neurons with an exponential-decay kernel, we display 100 randomly chosen neurons from default reservoir and tuned reservoir in Figure \ref{fig:reservoirResponse_pulse}.

We see that there are only two types of neuron activities in the default reservoir, whose dynamics are visualized in the left panel of Figure \ref{fig:reservoirResponse_pulse}. These two types of activities are (i) the ON-neurons fired at the time 0.5 seconds and (ii) the OFF-neurons fired at the time 1.5 second. On the other hand, the neuron activities produced by the tuned reservoir as shown in the right panel of Figure \ref{fig:reservoirResponse_pulse} are much more versatile. The highly versatile neuron responses produced by the tuned reservoir are usually favored for tasks such as regression and pattern classification. Intuitively, diverse neuron responses are more linearly separable. The versatility of reservoir responses exhibited by the tuned reservoir is also what one expects when conducting a similar \texttt{Pulse} experiment on a digital computer (c.f. \citep[Figure 3.5 C]{Enel2014}).

\subsection{The \texttt{Pulse-Chirp} experiment}

\begin{figure}[H]
        \centering
            \includegraphics[width=0.9 \textwidth]{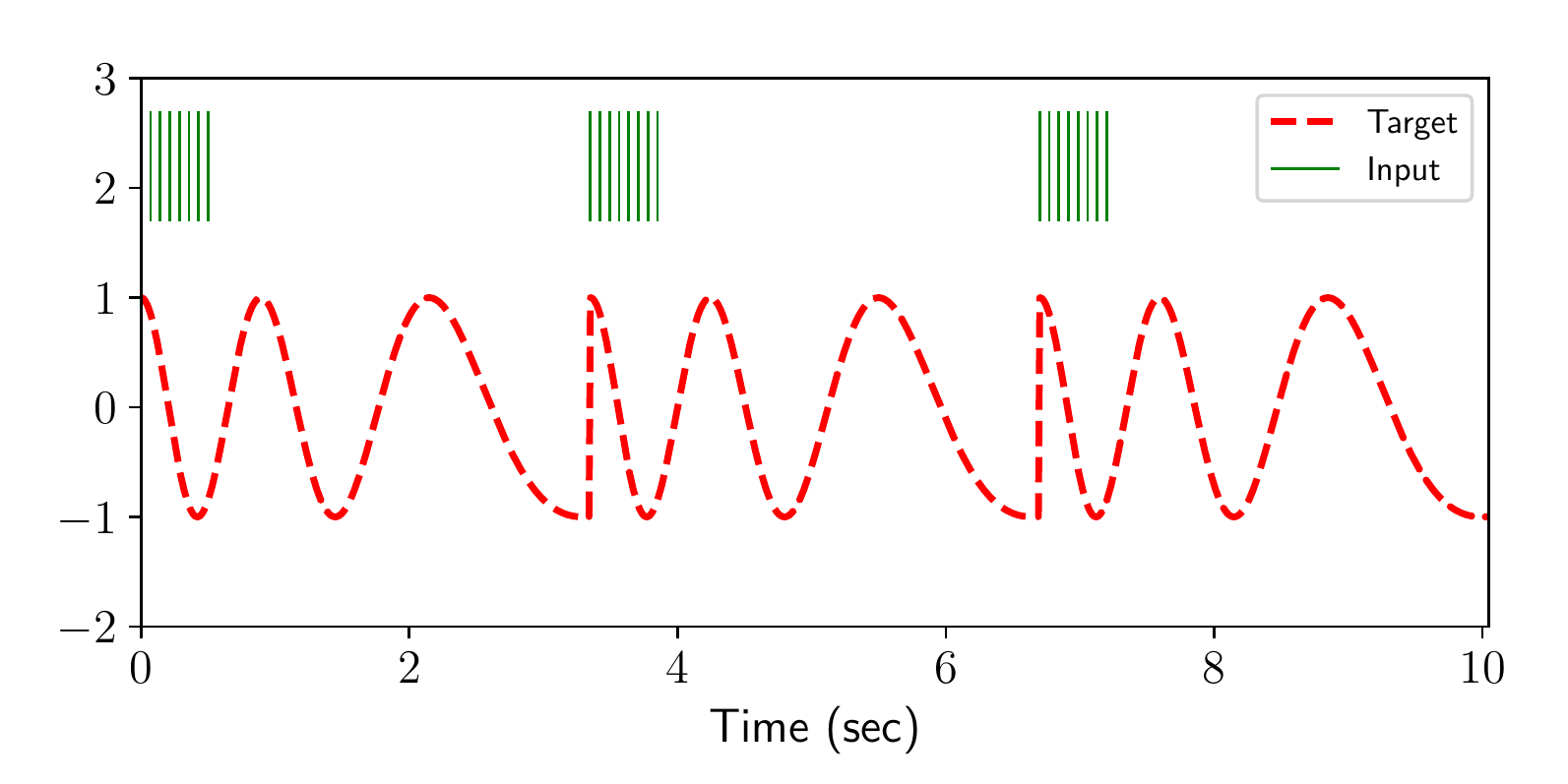}
     \caption{Visualization of three repetitions of input pulses and their target chirp signals. The pipeline of the \texttt{Pulse-Chirp} experiment is to (i) drive a reservoir with the input spike train (green vertical bars) and (ii) linearly map reservoir responses to the target chirp signal (red dashed line). For the linear map to work, the reservoir responses need to attain a certain length of memory. \label{fig:thesis_regression_input_output.pdf}}    
    \end{figure}

To further test the short-term memory of the default and tuned reservoirs, we conducted a \texttt{Pulse-Chirp} experiment similar to the one used in \citep{He2019}, whose presentation we follow here. The goal of this regression task is to learn an input-output map, where the input is a sequence of pulses with short widths separated by long periods of silence; the output is a chirp signal, whose oscillation frequencies are adapting over time. Since the values of the target chirp signal depend on the past values, the input-output map can only be successfully learned if the reservoir responses preserve some information about the input history. Three repetitions of such pulses (green vertical bars) and their corresponding 3 repetitions of target chirp signals (red curve) are illustrated in Figure \ref{fig:thesis_regression_input_output.pdf}. 

\begin{figure}[H]
        \centering
            \includegraphics[width=\textwidth]{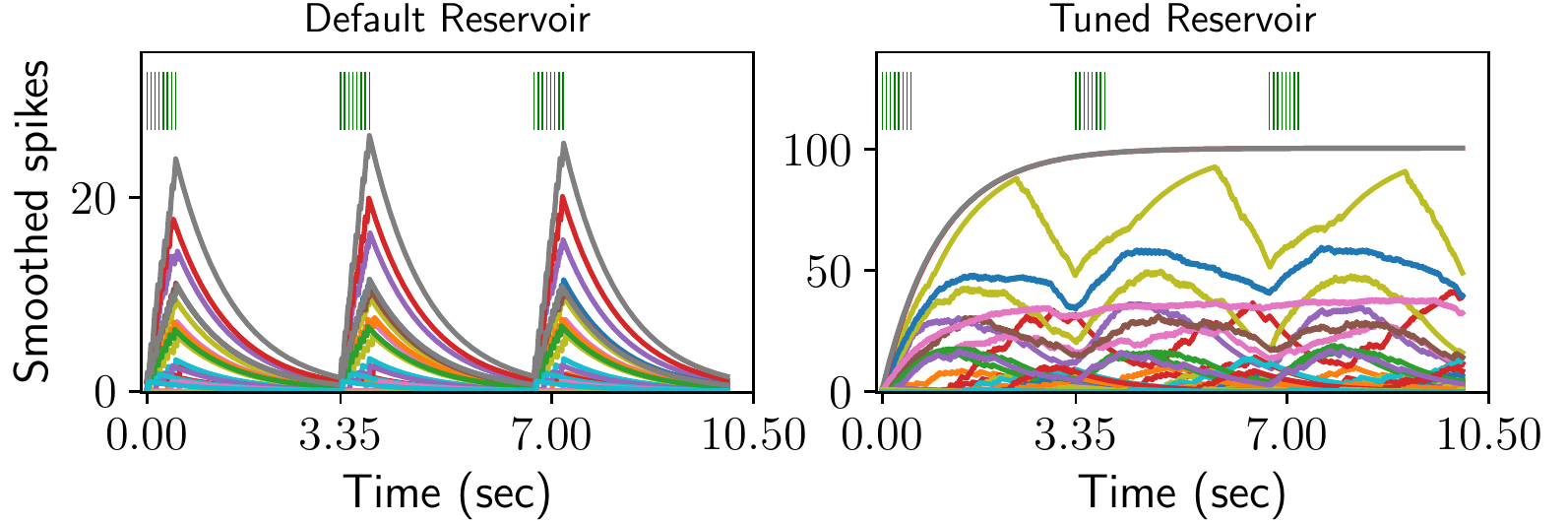}
     \caption{Visualization of the reservoir responses of a sequence of short pulses (0.5 seconds) gapped by long periods of silence (3 seconds). The sequence of pulses is visually illustrated with green vertical bars. For each of the default and tuned reservoir, we randomly choose 100 neurons and plot their neuronal responses (exponentially smoothed spikes) against time. Left: responses of the default reservoir. Right: responses of the tuned reservoir. \label{fig:reservoirResponse_regression}}    
    \end{figure}

In our experiment, the lasting time for each pulse block is 0.5 seconds and the gap between two pulse blocks is 2.85 seconds\footnote{We use this peculiar ``2.85 seconds'' due to a technical issue we encountered for this experiment. As Dynap-se disallows large gaps between two consecutive spikes, to introduce long silence time for this experiment, we employ a work-a-round solution: during the silent period, we send some ``pseudo spikes'' to on-chip neurons that are not used throughout the experiment. The list of ``pseudo spikes'' was converted from a linearly increasing continuous signal. Although this continuous signal lasts for 3 seconds, the spike train converted from it happens to last 2.85 seconds due to thresholding effect of the analog-to-spike conversion mechanism.} We repeated this input pattern for 30 times, resulting an input signal for Dynapse that lasts for $30 \times 3.35 = 100.5$ seconds. After 30 repetitions of pulses were sent, the responses of default and tuned reservoir neurons were collected. The collected spike trains were then smoothed with an exponential-decay kernel. Figure \ref{fig:reservoirResponse_regression} shows the responses of default and tuned reservoir when driven by the input spikes. Similar to what we have seen in Figure \ref{fig:reservoirResponse_pulse}, we observe that the reservoir responses from the tuned reservoir (right panel of Figure \ref{fig:reservoirResponse_regression}) is much more diverse than those from the default reservoir (left panel of Figure \ref{fig:reservoirResponse_regression}) when driven by the sequence of pulses.  
        
\begin{figure}[H]
        \centering
            \includegraphics[width=\textwidth]{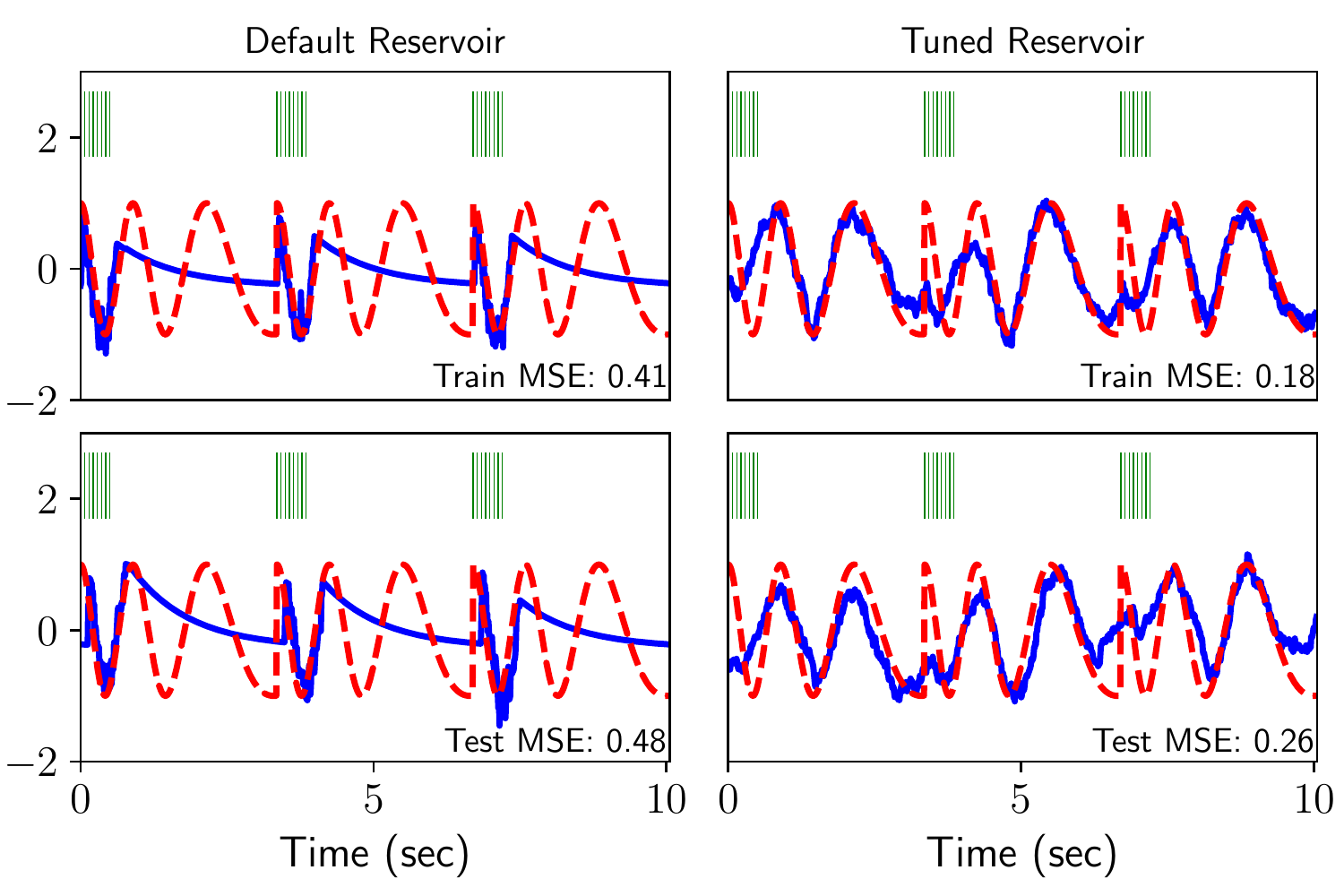}
     \caption{Training and testing results of the \texttt{Pulse-Chirp} regression task using the default and the tuned reservoir. Figures in the first column are training (first row) and testing (second row) results of a default reservoir; in a similar layout, figures in the second column are training and testing results of a tuned reservoir. The input sequences of spikes are illustrated with Green vertical bars; the target chirp signal is shown in red, and the predictions of the target chirp signal (linearly read out from reservoir) are in blue. Numbers inset are the mean square errors (MSEs) for training or testing datasets. \label{fig:result_regression}}    
    \end{figure}

So far, the experiment is similar to what we have done in the previously introduced \texttt{Pulse} experiment. Different from the \texttt{Pulse} experiment, however, we carried out a regression for this experiment, where the argument of the regression is the reservoir responses driven by these 30 repetitions of pulses and the target is 30 repetitions of chirp signals, whose oscillating frequencies vary with respect to time. To do so, we split the harvested reservoir responses into a training dataset and a testing dataset. The training dataset contains reservoir responses corresponding to the first 24 repetitions of input pulses and the test dataset contains the rest of the responses. A ridge regression was performed to map the reservoir responses from the training dataset to its corresponding target. We then submitted the training and testing reservoir responses for the same linear transformation, which is specified by ridge regression coefficients estimated using the training data. A portion of the training results and testing results (10 seconds each) for both reservoirs are shown in Figure \ref{fig:result_regression}. 

By comparing the left and right column of Figure \ref{fig:result_regression}, we see that the linear readout applied to the default reservoir neuron responses failed to replicate the time-adapting oscillation behavior of the target chirp signal (left panel), whereas the tuned reservoir solved the same task with much lower mean square error (right panel). This indicates that the memory length possessed by the tuned reservoir favorably outperforms the default one.

\subsection{The \texttt{Ramp + Sine} experiment}

\begin{figure}[H]
        \centering
        \begin{subfigure}[b]{0.5 \textwidth}   
            \centering 
            \includegraphics[width=\textwidth]{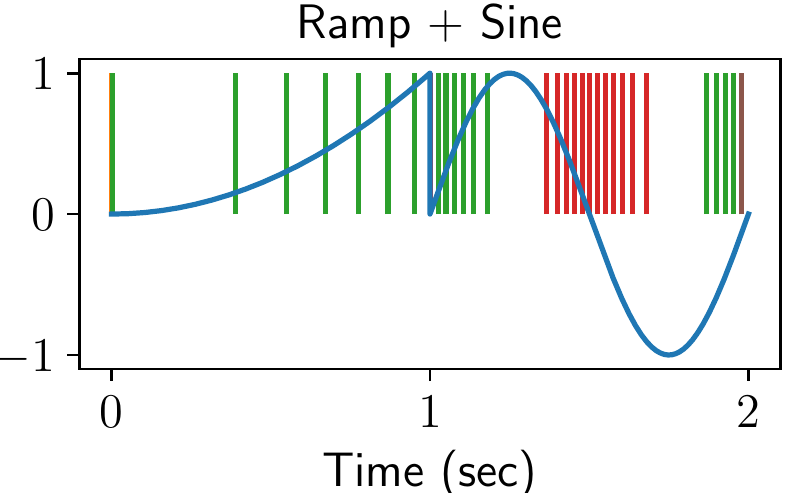}
        \end{subfigure}
        \hspace{-0.6em}
        \begin{subfigure}[b]{0.5 \textwidth}   
            \centering 
            \includegraphics[width=\textwidth]{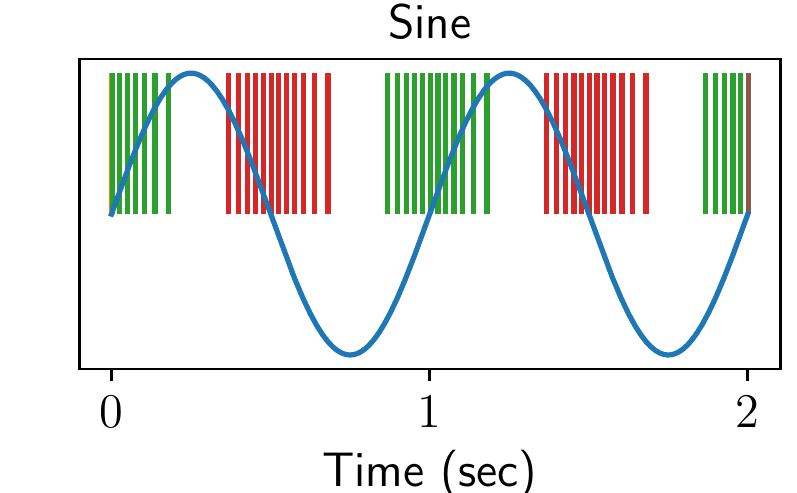}
        \end{subfigure}
        \caption{The \texttt{Ramp + Sine} and \texttt{Sine} signals and their respective converted input spike trains. The green vertical bars indicate the spikes which are assumed to be  generated by an excitatory input neuron and the red vertical bar indicate the spikes which are assumed to be generated by an inhibitory input neuron. Left panel: the converted spike train from the \texttt{Ramp + Sine} signal. Right panel: the converted spike train from the \texttt{Sine}. \label{fig:patterns}}
        
    \end{figure}

In this experiment, we aim to compare the performances of default and tuned reservoirs under a classification task. Our goal is to classify the temporal signal \texttt{Ramp+Sine} and \texttt{Sine}, which is depicted in Figure \ref{fig:patterns}. These two patterns are specifically designed such that the second half of the \texttt{Ramp + Sine} signal is the same as the second half of \texttt{Sine} signal. To correctly distinguish these patterns, the spiking neural network needs to maintain its memory when the temporal input proceeds into the second half of the patterns. To start the experiment, we converted the continuous ramp or sine signals into spike train as the input data for Dynap-se. The resulting input spike data is shown in the second row of Figure \ref{fig:patterns}, where the spikes in green are assumed to be generated by an excitatory neuron and the spikes in red are assumed to be generated by an inhibitory neuron.

In Figure \ref{fig:patterns}, each signal lasts for 2 seconds, and so do their corresponding spike trains. When performing the experiment, we sent 5 repetitions of each pattern into the Dynap-se, so that a single experiment lasts for $2 \times 5 \times 2 = 20$ seconds. For each pattern, we used the first two segments for the washout purpose and we only collected the responding spikes starting from the 3rd repetition of each pattern. We repeated the above process twice, once using the default reservoir and once using the tuned reservoir, to harvest their respective reservoir responses. We have appended the neural responses of the default and tuned reservoir driven by the \texttt{Ramp + Sine} pattern to the Figure \ref{fig:thesis_classification_visualization} in the Appendix.

\begin{figure}[H]
        \centering
            \includegraphics[width=\textwidth]{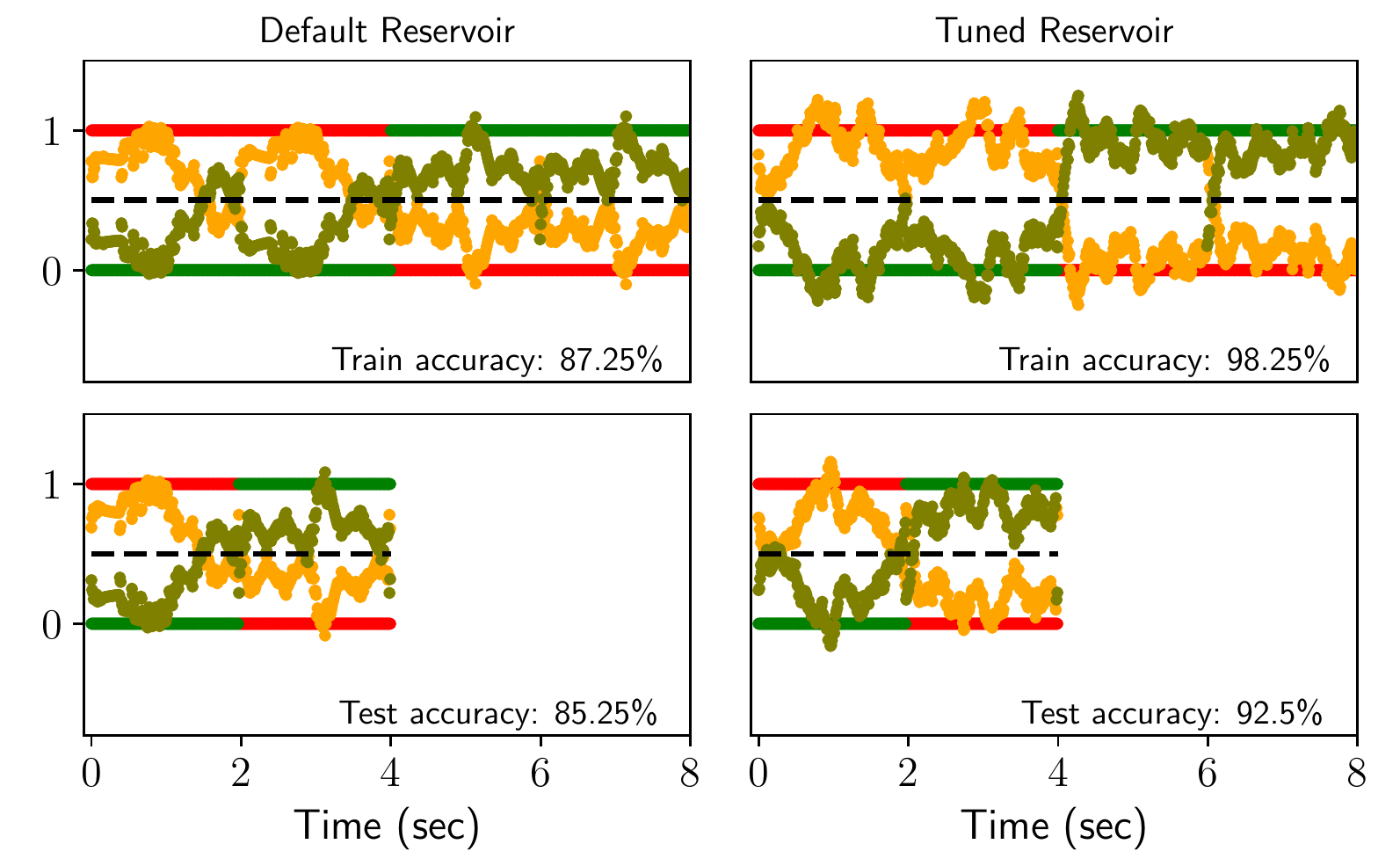}
     \caption{ Training and testing results of the \texttt{Sine-Ramp} classification task using the default and the tuned reservoir. Figures in the first column are training (first row) and testing (second row) results of a default reservoir; in a similar layout, figures in the second column are training and testing results of a tuned reservoir.  The thick red line and the thick green line at the $y$-axis 1 or 0 represent the regression targets for \texttt{Ramp+Sine} and \texttt{Sine}, respectively. The orange dots represent the predicted score for \texttt{Ramp+Sine}, and the green dots represent the predicted score for \texttt{Sine}. Numbers inset are predicted accuracies for training or testing datasets, where the accuracy is defined as the ratio of the number of corrected predicted bins (each bin lasts for 0.01 seconds). \label{fig:reservoirResponse_classification}}    
    \end{figure}

With the collected spikes from the reservoir, we performed spike data post-processing with an exponential-decay kernel as we have done before in the regression task. Next, we splitted the harvested reservoir responses into a training dataset and a testing dataset. The training dataset contains reservoir responses corresponding to two repetitions of each input pattern and the testing dataset consists of the rest of the reservoir responses. With the training data, we performed a ridge regression to extract the features of two patterns. The input argument for the ridge regression is the training dataset of the smoothed spike trains. The regression target is a matrix whose columns are one-hot encoded indication of the signal, where the column vector $[1, 0]^\top$ is the target for \texttt{Ramp+Sine} pattern and $[0, 1]^\top$ is the target for \texttt{Sine} pattern. Ridge regression coefficients are calculated based on the training dataset and its target. For prediction, we linearly transformed the training and testing reservoir responses using the learned coefficients. In Figure \ref{fig:reservoirResponse_classification} we display the classification results for the signal. 
  
By comparing the left and right panel of Figure \ref{fig:reservoirResponse_classification}, we see that the input signals processed by the tuned reservoir (right panel) achieved much better classification performances than those processed by the untuned reservoir (left panel).

\chapter{Slowing down Neuronal Dynamics by Modifying the Reservoir Topology  \label{chap:resTransfer}}

In Chapter \ref{chap:paramtuning} we introduced heuristic techniques which slow down neural dynamics by modifying properties of individual neurons. Instead of working on the single neuron level, we can also directly modify the global properties of a population of neurons. In this chapter, we introduce such a global method named \emph{Reservoir Transfer}. The method maps the desired dynamic properties of a RNN whose dynamics is well-tuned on a digital computer to an on-chip spiking RNN.

A version of this chapter has been published as \citep{He2019}.  T. Liu contributed to the paper as the second author by (i) creating a dataset for the purpose of training the on-chip reservoir (to be discussed in Section \ref{sec:TrainOnChipNet} of this thesis) and (ii) using the on-chip reservoir to carry out an ECG heartbeat abnormality experiment (to be discussed in Section \ref{sec:ECGexperiment}). In the following section, we present the Reservoir Transfer method sometimes using the wording of \citep{He2019}.

\section{Reservoir Transfer}  \label{sec:ReservoirTransfer}

The idea of the Reservoir Transfer method is based on the insight that the dynamics of a population of neurons can be slower than those of individual neurons. For artificial recurrent neural networks on a conventional computer, slow dynamics can be attained by properly choosing the global network parameters, e.g., spectral radius of the recurrent connectivity matrix. However, setting these parameters on Dynap-se based recurrent neural network is impractical due to its low numerical precision. To address this issue, \citet{He2019} proposed to ``transfer'' the dynamic properties of a well-performing RNN of artificial neurons (the ``teacher network'') to on-chip RNNs of leaky integrate-and-fire neurons (the ``student network'). In this section, we introduce the teacher network, the student network, and the transfer mechanism. 

\subsection{The teacher network}
We first define a teacher network operating on a conventional computer, whose dynamics we wish to ``mirror'' to a student network. The teacher network we use is an Echo State Network with leaky integrator neurons \citep{Jaeger2001}. When driven by a sequence of $m$-dimensional input signal $\mathbf{u}(t)$ at the time $t$, the evolution of the $N$-dimensional continuous-time state vector $\mathbf{x}(t)$ of the network is given by

\begin{equation} \label{eq:ESN}
    \dot{\mathbf{x}}(t) =-\lambda_x \mathbf{x}(t) + \tanh( \mathbf{W}^{\text{in}}\mathbf{u}(t) + \mathbf{W} \mathbf{x}(t)),
\end{equation}
where $\lambda_x$ is the leaking rate, $\mathbf{W}^{\text{in}} \in \mathbb{R}^{N \times m} $ and $\mathbf{W} \in \mathbb{R}^{N \times N}$ are input and recurrent weights. In a reservoir computing paradigm, the input weight matrix $\mathbf{W}^{\text{in}}$ and recurrent weight matrix $\mathbf{W}$ are randomly generated according to some global parameters such as the scaling factor of $\mathbf{W}^{\text{in}}$ and the spectral radius of $\mathbf{W}$ \citep{Lukosevicius2012}.  

\subsection{The student network}

We now present the student Spiking Neural Network (SNN), which is the RNN of integrate-and-fire (LIF) neurons introduced in Equation \ref{eq:RNN_LIF} of Section \ref{sec:RNN_LIF}. Note that this student network is slightly different from the one used in the original reservoir transfer paper \citep{He2019} in that the time constants are placed at different locations. Since the reservoir transfer method will not be influenced by these changes, here we use the RNN with LIF neurons introduced in Equation \ref{eq:RNN_LIF} for consistency. 

Recall from Equation \ref{eq:RNN_LIF} that, when driven by an $m$-dimensional input signal $\mathbf{u}(t)$ at the time $t$, the dynamics of a recurrent neural network of LIF neurons can be described by 

 \begin{equation} \label{eq:RNN_LIF_repeat}
\begin{aligned}
  \tau_v \dot{\mathbf{v}} & =- \mathbf{v}(t)  + \mathbf{\hat{W}}^{\text{in}} \mathbf{u} (t) +  \mathbf{\hat{W}} \mathbf{r}(t)  + \mathbf{I}_0 - \theta \mathbf{s}(t),  \\
 \tau_r \dot{\mathbf{r}} & = - \mathbf{r}(t) + \mathbf{s} (t),
 \end{aligned}
 \end{equation}
where $\mathbf{v}, \mathbf{s}$, and $\mathbf{r}$ are $N$-dimensional vectors whose $i$-th entries are denoted by $v_i, s_i$, and $r_i$, respectively: $v_i$ is the membrane potential of the $i$-th neuron, $s_i(t)=\sum_{t_f^i}\delta(t-t_f^i)$ is the neuron's output spike train with spike times $t_f^i$ together with a Dirac delta function $\delta (\cdot)$, and $r_i$ is the exponentially decaying synaptic currents triggered by $s_i$; $\mathbf{I}_0$ is a vector whose entries are all $I_0$, a constant current set near or at the rheobase (threshold to spiking) value \citep{Nicola2017}; the matrices $\hat{\mathbf{W}}^{\text{in}} \in \mathbb{R}^{N \times m}$ and $\hat{\mathbf{W}}\in \mathbb{R}^{N \times N}$ are input weights and recurrent weights of the student SNN.  Note that the teacher network and the student network have the same number of neurons.

\subsection{Transfer dynamics of the teacher network to the student network}

Since the reservoir dimension $N$ are usually much higher than the input signal dimension $m$, the state vectors $\mathbf{x}(t)$ of the teacher ESN in Equation \ref{eq:ESN} can be seen as high-dimensional temporal features of the input signal $\mathbf{u}(t)$. To transfer the dynamic properties of the teacher ESN to the student SNN, we inject these features $\mathbf{x}(t)$ of the teacher ESN element-wisely into the corresponding student SNN, replacing the recurrent inputs $\hat{\mathbf{W}}\mathbf{r}(t)$ in Equation \ref{eq:RNN_LIF_repeat}. The resulting dynamics of the SNN can be described by

 \begin{equation} \label{eq:v}
\begin{aligned}
  \tau_v \dot{\mathbf{v}}_x & =- \mathbf{v}_x(t)  + \hat{\mathbf{W}}^{\text{in}} \textbf{u} (t) + \mathbf{x}(t) + \mathbf{I}_0 - \theta  \mathbf{s}_x(t),  \\
 \tau_r \dot{\mathbf{r}}_x & = - \mathbf{r}_x (t) + \mathbf{s}_x (t).
 \end{aligned}
 \end{equation}

In Equation \ref{eq:v}, the dynamics of the student SNN are sustained with the help of $\mathbf{x}(t)$. Ideally, however, we would like the same dynamics $\mathbf{v}_x(t)$ of the student SNN to be sustained without manually injecting those $\mathbf{x}(t)$. To this end, we would like to choose a $\hat{\mathbf{W}}$, such that when both networks are driven by the same input signal $\mathbf{u}(t)$, the dynamics of two networks are similar in the sense that $\hat{\mathbf{W}}\mathbf{r}_x(t) \approx \mathbf{x}(t)$ for all $t$. To estimate such $\hat{\mathbf{W}}$, however, it is practically infeasible to take all kinds of input signal $\mathbf{u}(t)$ and all continuous-valued time $t$ into account. For this reason, we resorted to a more modest goal: we fixed $\mathbf{u}(t)$ to be a white noise signal and use it to drive the teacher and student networks. We then compute $\hat{\mathbf{W}}$ by letting 

\begin{align}\label{eq:regression}
\hat{\mathbf{W}} \coloneqq \argmin_{\tilde{\mathbf{W}}} \sum_{t_k} \| \tilde{\mathbf{W}}\mathbf{r}_x(t_k)-\mathbf{x}(t_k) \|_2^2,
\end{align}
where $t_k$ are some discrete time samples and $\mathbf{r}_x$ and $\mathbf{x}$ are those reservoir responses when driven by the fixed white noise signal $\mathbf{u}_t$. The $\hat{\mathbf{W}}$ in Equation \ref{eq:regression} can be solved via a linear regression. The solved $\hat{\mathbf{W}}$ can then be used as a reservoir in the student SNN.

We note that the reservoir transfer method is not limited to the choice of neuron model used in the student spiking neural network. In Equation \ref{eq:RNN_LIF_repeat} we used RNN of LIF neurons for the convenience of presentation. Other types of neuron models can be straightforwardly used in the reservoir transfer paradigm, too. Indeed, the neurons equipped on Dynap-se are based on the AdEx model \citep{Brette2005}, which is a generalization of the leaky integrate-and-fire model.

\section{Training on-chip reservoir} \label{sec:TrainOnChipNet}

We employed the reservoir transfer method to train a reservoir of 768 neurons (3 cores) on Dynap-se. The training procedure is outlined as follows. We first created a leaky ESN of equal size in the Brian2 simulator \citep{Brette2009} as a teacher reservoir. We sent a white noise input signal $\mathbf{u}(t)$ to the teacher ESN to harvest its reservoir responses. These responses are then converted to spike trains and are sent to the student network on Dynap-se, whose parameters have already been tuned according to the heuristic techniques discussed in Chapter \ref{chap:paramtuning} in advance. After the output spike trains from the hardware neurons are recorded, we smoothed both the input and output spike trains by an exponential decay kernel to get $\mathbf{x}(t)$ and $\mathbf{r}(t)$, respectively. Instead of using the standard linear regression to solve Equation \ref{eq:regression}, we employed a ternarized linear regression \citep{Zhu2017} to compute the weight matrix $\hat{\mathbf{W}}_{\text{ternary}}$ of ternary precision. That is, the values of the matrix $\hat{\mathbf{W}}_{\text{ternary}}$ are either -1, 0, or 1, corresponding to inhibitory synapses, no connection, and excitatory synapses. Ternarized linear regression was used here because our Dynap-se hardware does not support full-precision recurrent connectivities. The learned ternary connectivity matrix was then written into a \texttt{.txt} file and loaded into Dynap-se as the trained reservoir. When writing the learned topology into Dynap-se readable format, we assumed that all synapses are slow synapses according to the recommendation of Heuristic Technique 2 introduced in Chapter \ref{chap:paramtuning}. 

One advantage of reservoir transfer method for Dynap-se hardware is that the method does not require exact values of the network state variables such as membrane potentials and currents, which are unobservable and varying across individual neurons on Dynap-se. The neurons do not have to share the same parameter value as long as their collective response to the input current $\mathbf{x}(t)$ contains enough information to linearly decode $\mathbf{x}(t)$. Moreover, learning is needed only once using a white noise signal, afterward, the connection weights can stay fixed. Hence no online adaptation on hardware is needed.

We would like to briefly remark the similarities and differences of the reservoir transfer method and the pre-trained DNN method introduced in section \ref{sec:deepLearningNeuromorphic}. Both methods map artificial neural networks to their counterparts on neuromorphic devices. However, the objectives of reservoir transfer method and the pre-trained DNN are quite different. The pre-trained DNN approach first learns parameters based on a particular task (e.g., image classification) and maps the learned parameters to neuromorphic hardware such that the on-chip neural networks can solve the \emph{same} task. The reservoir method, however, is not optimized with respect to a particular task. Instead, the learning here aims to map characteristics of slow dynamics of the teacher ESN to the student SNN. On can say that the learned connection weights resulted from the reservoir transfer method are not task-customized but timescale-customized.

\section{ECG monitoring experiment} \label{sec:ECGexperiment}


\begin{figure}[htbp]
    \begin{centering}
        \includegraphics[width=1.0\columnwidth]{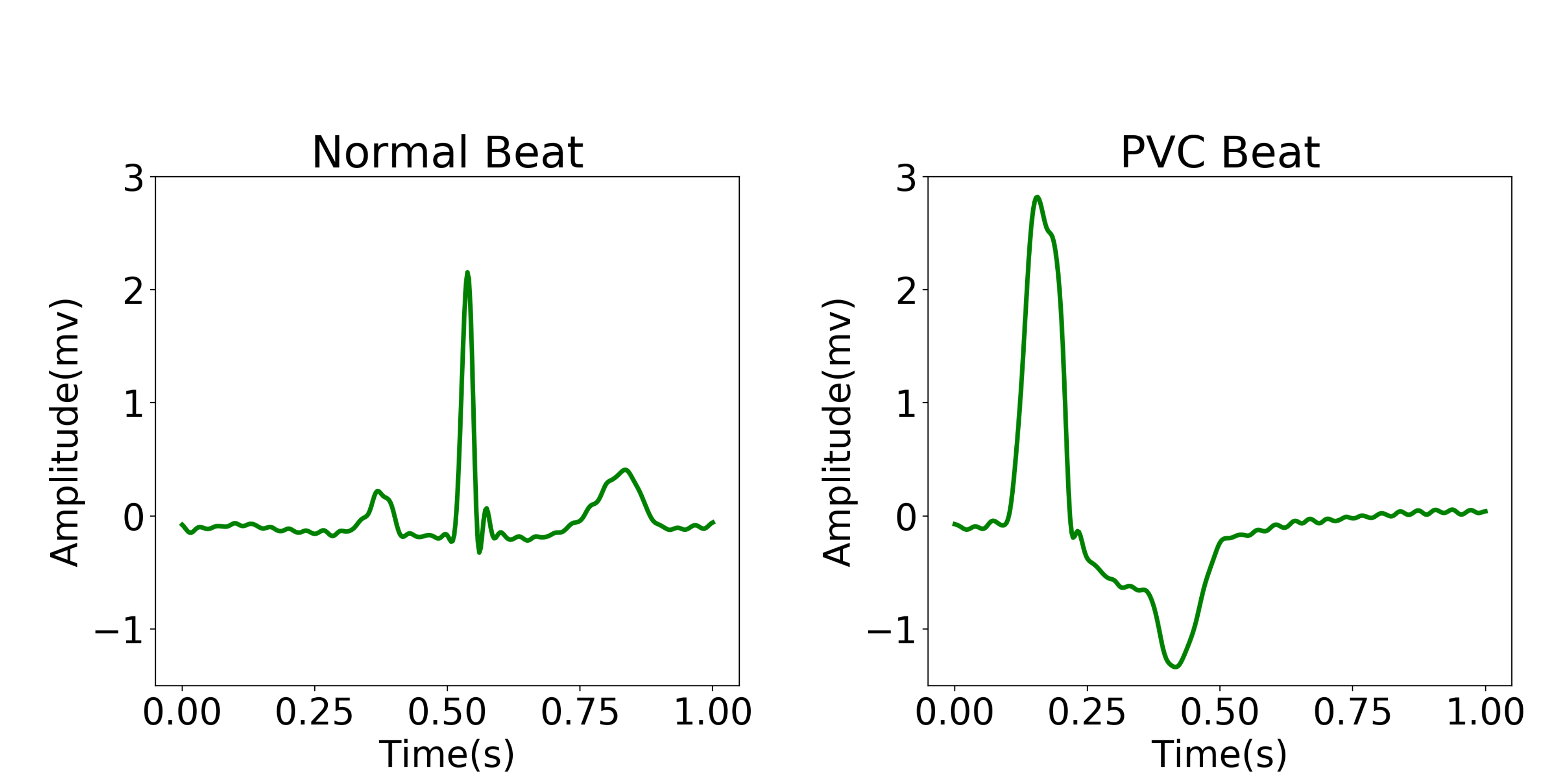}
        \par\end{centering}
    \caption{Two patterns of heartbeats in an ECG signal. Left panel: a normal heartbeat. Right panel: a PVC heart beat.}
    \label{fig:beatSample}
\end{figure}

To verify that the transfer learning method yields a functional physical spiking reservoir, we conducted ECG signal classification experiments using the learned reservoir on Dynap-se. The experiment aims to detect Premature Ventricular Contractions (PVCs), which are abnormal heartbeats initiated by the heart ventricles. Figure \ref{fig:beatSample} shows a normal heartbeat (left panel) and a PVC heartbeat (right panel). More concretely, we formulated the PVC detection task as a supervised temporal classification problem which demands a binary output signal $y(n)$ for each heartbeat indexed by $n$:

\begin{equation} \label{eq:target}
y(n) = \begin{cases}
    1  &  \text{if the $n$-th heartbeat is a PVC},\\
    0  &  \text{otherwise}.
  \end{cases}
\end{equation}

We used the MIT-BIH ECG arrhythmia database \cite{Goldberger2000} in this experiment. The database provides 48 half-hour excerpts of two-channel ambulatory ECG recording files, obtained from 47 different patients. The recordings were digitized with a sampling frequency of 360 Hz and acquired with 11-bit resolution over a 10mV range. Each record was annotated by two or more cardiologists independently, both in timing information and beat classification. In this work, we used recordings from file \#106, \#119, \#200, \#201, \#203, \#223, and \#233. We aim to train a classifier for each subject, such that the trained classifier can distinguish the normal and abnormal heartbeats of the corresponding subject. To this end, we used the annotation file of each subject to train and evaluate the classifier. More concretely, we used an interval of 10 minutes of the recording signal from each subject for training and the next 5 minutes for testing. We carried out numerical experiments using the following routine.

\begin{enumerate}
\item \emph{ECG pre-processing}: we removed the baseline drift from an ECG signal by applying a high-pass Butterworth filter and then normalized the signal into the numerical range [0,1]. 

\item \emph{Signal-to-spike conversion}: we placed a spike at a time index if the increase/decrease of the ECG signal relative to its value at the previous spike time surpassed a threshold of numerical value 0.1. 
\item \emph{Reservoir response harvesting}: we sent ECG-converted spike trains into Dynap-se to harvest the reservoir responses, which were in the form of spike trains. 

\item \emph{Spike-to-signal conversion}: on a digital computer, we smoothed the spike trains collected from the physical reservoir to continuous-valued time-series by an exponential decay kernel with a decay time constant, which is a hyperparameter for each individual subject. 

\item \emph{Classifier training}: the training of the classifier amounted to solving a linear regression problem, where the input for linear regression was the smoothed reservoir responses and the target output was a $\{0,1\}$-valued binary signal indicating the correct labels of heartbeats. To derive a stable linear regression solution, we used a ridge regression in practice. Dividing the reservoir responses and target signal into five segments, we used a five-fold cross-validation scheme to optimize the learning parameters, which include a regularization coefficient for ridge regression and a binarization threshold to round the predicted labels to 0 and 1. 

\item \emph{Test result evaluation}: with a testing ECG time-series, we repeated the above procedure to procure its smoothed reservoir responses and then readout the predicted labels with learned weights. We used the following familiar metrics to evaluate the binary classification performance: accuracy, sensitivity, precision, and F1-score. Concretely, letting TP denote the number of true positive predictions (abnormal heartbeats correctly identified as abnormal), FP denote the number of false positive (normal heartbeats incorrectly identified as abnormal heartbeats), TP denote the number of true negative predictions (normal heartbeats correctly identified as normal), and FN denote the number of false negative predictions (abnormal heartbeats incorrectly identified as normal), the metrics accuracy, sensitivity, precision, and F1-score are defined as follows: Accuracy = (TP + TN)/(TP + TN + FP + FN), Sensitivity = TP/(TP + FN), Precision = TP/(TP + FP), F1-score = 2TP/(2TP + FP + FN).

\end{enumerate}

A comparison of classification accuracy on testing data between the low-precision spiking reservoir and the digitally simulated, high-precision reservoir baseline is provided in Table \ref{tab:result}. The high-precision reservoir baseline is a standard ESN whose parameters set as leakage rate = 0.99, spectral radius= 0.9, and regression parameter = 1e-6.

\begin{table}[H] 
\centering
\caption{PVC detection results on testing data \label{tab:ECGresult}}

\begin{tabular}{ r c c c c c} \toprule
 &  & \multicolumn{4}{c}{Performance Metrics}\\ \cline{3-6}
subject number   &  classifier & Accuracy &Sensitivity &Precision &F1   \\ \midrule
 \multirow{2}{*}{subject \#106}     & Standard ESN           &  98.75 \% &  97.22 \%&  97.22 \%    &  97.22 \% \\
                         & Dynap-se reservoir      & 91.30 \% & 88.89  \%& 76.19 \%    & 82.05 \% \\ \midrule 
 \multirow{2}{*}{subject \#119}     & Standard ESN           & 99.70 \% & 100 \%    &99.10 \%& 99.55 \% \\
                         & Dynap-se reservoir      & 97.87 \% & 100   \%      & 94.07 \%  & 96.94 \% \\ \midrule 
 \multirow{2}{*}{subject \#200}     & Standard ESN           & 99.07 \% & 98.24  \% & 99.40 \% & 98.82 \% \\
                         & Dynap-se reservoir      & 95.80 \% & 93.53 \%& 95.78 \% & 94.64 \% \\ \midrule 
 \multirow{2}{*}{subject \#201}     & Standard ESN           & 99.24 \% & 100  \% & 97.18 \% & 98.57 \% \\
                         & Dynap-se reservoir      & 97.74 \%& 95.71 \%& 95.71 \%& 95.71 \% \\ \midrule 
 \multirow{2}{*}{subject \#203}     & Standard ESN           & 98.14 \%& 100  \% & 90.32 \% & 94.92 \% \\
                         & Dynap-se reservoir      & 89.28  \%& 79.38 \% & 70.64  \% & 74.76 \%\\ \midrule 
 \multirow{2}{*}{subject \#223}     & Standard ESN           & 99.07 \%& 99.05  \% & 98.11 \% & 98.58 \%\\
                         & Dynap-se reservoir      & 90.53 \%& 76.15 \%& 84.69  \%& 80.19\% \\ \midrule 
  \multirow{2}{*}{subject \#233}     & Standard ESN           & 99.78 \%& 100  \%& 99.21  \% & 99.60 \%\\
                         & Dynap-se reservoir      & 97.46 \%& 93.01 \%& 97.79 \% & 95.34 \%\\ \bottomrule
 \end{tabular} 
\label{tab:result}
\vspace{10pt}
\end{table}

From Table \ref{tab:ECGresult}, we see that the computational performance of the on-chip neural networks favorably approximate that of ESNs on conventional computers.

\chapter{Conclusion} \label{chap:conclusion}
In this thesis, we reported our attempts to realize slow reservoir dynamics on a type of analog neuromorphic hardware named Dynap-se. We empirically demonstrated that by harnessing slow dynamics, spiking neural networks deployed on analog neuromorphic hardware can gain non-trivial performance boosts for real-time signal processing tasks. We now summarize the contributions of this thesis. 

In Chapter \ref{chap:neuromorphicComp}, we outlined a general pipeline for conducting experiments with Dynap-se board.  This pipeline can be used as a primer for practitioners who wish to conduct numerical experiments on Dynap-se. In Chapter \ref{chap:paramtuning} we proposed two heuristics methods for slowing down the dynamics of on-chip neural networks. Since these two techniques operate locally at the neuron level, they
can be conveniently applied to Dynap-se for different tasks. In Chapter \ref{chap:resTransfer} we introduced the reservoir transfer paradigm, which  ``mirrors'' well-tuned dynamics of an artificial neural network to an on-chip spiking neural network. For the reservoir transfer paradigm, the main contribution of the thesis was on the experiment side, for which we have tested the effectiveness of the transferred reservoir using ECG datasets collected from 7 subjects. 

This thesis has a few limitations. An important one is that we need more thorough investigations on the separate roles played by the parameter tuning heuristics and the reservoir transfer method. As pointed out in Chapter \ref{chap:resTransfer}, when training the on-chip reservoir (Section \ref{sec:TrainOnChipNet}) and when conducting the ECG experiments (Section \ref{sec:ECGexperiment}), we used \emph{heuristically tuned} parameters. That is, the reservoir transfer pipeline operated with the help of the tuned parameters. By doing so, two important issues remain unclear: (i) Will reservoir transfer work using the untuned set of parameters? (ii) How well can the tuned (yet untrained) reservoir perform under the ECG experiments? To address these questions, more controlled experiments are needed. A second limitation of this thesis is that the experiments we have reported in Chapter \ref{chap:paramtuning} and Chapter \ref{chap:resTransfer} are based on on-chip reservoir responses driven by single trials of input spike trains. Recall that, for example, when conducting the \texttt{Pulse-Chirp} experiment in Subsection \ref{subsec:pulse-chirp}, the training and testing data were two separate segments of reservoir responses driven by \emph{a single trial} of input spike train. This approach, however, fails to take trial-to-trial variability of on-chip neural networks. Due to the stochasticity of analog circuits, regression weights estimated from reservoir responses driven by one trial of input spikes may perform well upon in-trial reservoir responses yet fail to generalize well to out-of-trial reservoir responses.
 
This thesis calls for a deeper investigation of the effects of timescales in spiking neural networks. In the future, we expect more algorithms that bring slow dynamics to spiking neural networks on analog neuromorphic hardware. Concretely, for future work, it will be worthwhile to formally validate/falsify the two heuristic techniques proposed in Chapter \ref{chap:paramtuning} by delving deep into the non-linear dynamics of DPI circuits. A second avenue for future research is to extend the existing reservoir transfer method. As pointed out in \citet{He2019}, instead of using a randomly created ESN for reservoir transfer, we plan to explore the effects of transferring \emph{trained} recurrent neural networks on neuromorphic hardware.

\appendix
\chapter{Parameters Values \label{chap:appendix}}

 \section{Default Parameters}

\begin{table}[H]
\begin{center}
\begin{tabular}{llll}\toprule
 & Parameter Names & Coarse Values & Fine Values     \\ \midrule 
 \multirow{11}{*}{Neuron Parameters} & IF\_AHTAU\_N          & 7 & 35         \\
&IF\_AHTHR\_N          & 7 & 1          \\
&IF\_AHW\_P            & 7 & 1          \\
&IF\_BUF\_P            & 3 & 80         \\
&IF\_CASC\_N           & 7 & 1          \\
&IF\_DC\_P             & 7 & 0          \\
&IF\_NMDA\_N           & 7 & 0          \\
&IF\_RFR\_N            & 4 & 60        \\
&IF\_TAU1\_N           & 7 & 130         \\
&IF\_TAU2\_N           & 0 & 100         \\
&IF\_THR\_N            & 7 & 130         \\
\multirow{14}{*}{Synapse Parameters} &NPDPIE\_TAU\_F\_P     & 4 & 36         \\
&NPDPIE\_TAU\_S\_P     & 5 & 38         \\
&NPDPIE\_THR\_F\_P     & 2 & 200        \\
&NPDPIE\_THR\_S\_P     & 2 & 200          \\
&NPDPII\_TAU\_F\_P     & 5 & 41        \\
&NPDPII\_TAU\_S\_P     & 5 & 41         \\
&NPDPII\_THR\_F\_P     & 0 & 150         \\
&NPDPII\_THR\_S\_P     & 7 & 150         \\
&PS\_WEIGHT\_EXC\_F\_N & 0 & 30        \\
&PS\_WEIGHT\_EXC\_S\_N & 0 & 100          \\
&PS\_WEIGHT\_INH\_F\_N & 0 & 100 \\
&PS\_WEIGHT\_INH\_S\_N & 0 & 114          \\
&PULSE\_PWLK\_P        & 2 & 112         \\
&R2R\_P                & 4 & 85    \\    \bottomrule
\end{tabular}
\caption{The default parameters of Dynap-se. These parameters can be configured by pressing ``set default spiking biases'' button on the GUI of Dynapse. \label{tab:defaultParams}}
\end{center}
\end{table}

 \section{Tuned Parameters}

\begin{table}[H]
\begin{center}
\begin{tabular}{llll}\toprule
 & Parameter Names & Coarse Values & Fine Values     \\ \midrule 
 \multirow{11}{*}{Neuron Parameters} & IF\_AHTAU\_N          & 7 & 35         \\
&IF\_AHTHR\_N          & 7 & 1          \\
&IF\_AHW\_P            & 7 & 1          \\
&IF\_BUF\_P            & 3 & 80         \\
&IF\_CASC\_N           & 7 & 1          \\
&IF\_DC\_P             & 7 & 0          \\
&IF\_NMDA\_N           & 7 & 0          \\
&IF\_RFR\_N            & 4 & 60        \\
&IF\_TAU1\_N           & 7 & 130         \\
&IF\_TAU2\_N           & 0 & 100         \\
&IF\_THR\_N            & 7 & 130         \\
\multirow{14}{*}{Synapse Parameters} &NPDPIE\_TAU\_F\_P     & \bf \textcolor{red} 7 & \bf \textcolor{red} 0         \\
&NPDPIE\_TAU\_S\_P     & 5 & 38         \\
&NPDPIE\_THR\_F\_P     & \bf \textcolor{red} 7 & \bf \textcolor{red} 0        \\
&NPDPIE\_THR\_S\_P     & 2 & 200          \\
&NPDPII\_TAU\_F\_P     & \bf \textcolor{red} 7 &\bf \textcolor{red}  0        \\
&NPDPII\_TAU\_S\_P     & 5 & 41         \\
&NPDPII\_THR\_F\_P     & \bf \textcolor{red} 7 & \bf \textcolor{red} 0         \\
&NPDPII\_THR\_S\_P     & 7 & 150         \\
&PS\_WEIGHT\_EXC\_F\_N & 0 & 30        \\
&PS\_WEIGHT\_EXC\_S\_N & 0 & 100          \\
&PS\_WEIGHT\_INH\_F\_N & 0 & 100 \\
&PS\_WEIGHT\_INH\_S\_N & 0 & 114          \\
&PULSE\_PWLK\_P        & 2 & 112         \\
&R2R\_P                & 4 & 85    \\    \bottomrule
\end{tabular}
\caption{The tuned parameters of Dynap-se. Compared to the default parameters in Table \ref{tab:defaultParams},  the modified ones are marked in red. \label{tab:tunedParams}}
\end{center}
\end{table}

 \section{Reservoir responses in \texttt{Ramp + Sine} experiment}

     \begin{figure}[H]
        \centering
            \includegraphics[width=0.8 \textwidth]{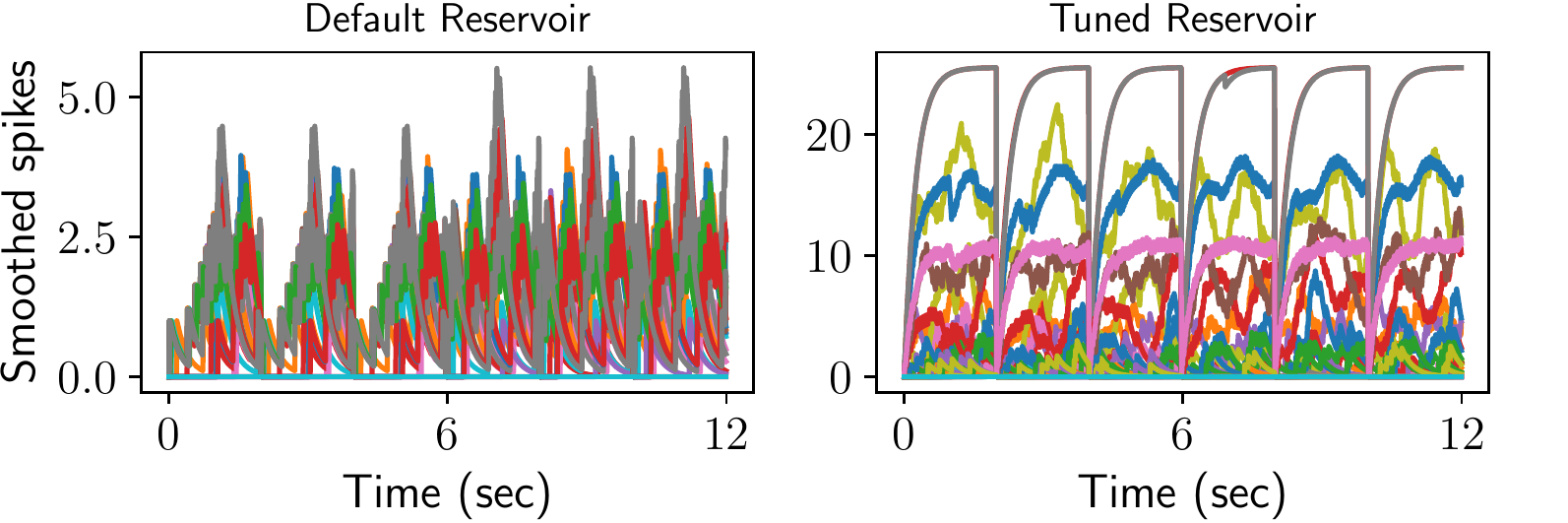}
     \caption{Visualization of the reservoir responses when driven by input signal from \texttt{Ramp + Sine} experiment. For each of the default and tuned reservoir, we randomly choose 100 neurons and plot their neuronal responses (exponentially smoothed spikes) against time. Left: responses of the default reservoir. Right: responses of the tuned reservoir. \label{fig:thesis_classification_visualization}}      
    \end{figure}

\backmatter

\bibliographystyle{plainnat}
\bibliography{/Users/liutianlin/Desktop/Academics/MINDS/neuromorphic/neuromorphic} 

\end{document}